\documentclass[letterpaper]{article} 
\usepackage{aaai24}  
\usepackage{times}  
\usepackage{helvet}  
\usepackage{courier}  
\usepackage[hyphens]{url}  
\usepackage{graphicx} 
\usepackage{natbib}  
\usepackage{caption} 
\usepackage{algorithm}
\usepackage{algorithmic}

\usepackage{subcaption}
\usepackage{amsmath}
\usepackage{amssymb}
\usepackage{mathtools}
\usepackage{amsthm}
\usepackage[capitalize,noabbrev]{cleveref}
\usepackage{multirow}
\usepackage{tabularx}
\usepackage{booktabs}
\usepackage{array}

\usepackage{newfloat}
\usepackage{listings}
\DeclareCaptionStyle{ruled}{labelfont=normalfont,labelsep=colon,strut=off} 
\floatstyle{ruled}
\newfloat{listing}{tb}{lst}{}
\floatname{listing}{Listing}
\title{HyperFast: Instant Classification for Tabular Data}
\author{%
  David Bonet\textsuperscript{\rm 1,2},\; Daniel Mas Montserrat\textsuperscript{\rm 1},\; Xavier Giró-i-Nieto\textsuperscript{\rm 3}\thanks{Work done prior to joining Amazon.},\; Alexander G. Ioannidis\textsuperscript{\rm 1} \\%
}
\affiliations{
  \textsuperscript{\rm 1}Stanford University, Stanford, CA, USA\\
  \textsuperscript{\rm 2}Universitat Polit\`{e}cnica de Catalunya, Barcelona, Spain\\
  \textsuperscript{\rm 3}Amazon, Barcelona, Spain \\
    ioannidis@stanford.edu
}

\begin{document}

\maketitle

\begin{abstract}
Training deep learning models and performing hyperparameter tuning can be computationally demanding and time-consuming. Meanwhile, traditional machine learning methods like gradient-boosting algorithms remain the preferred choice for most tabular data applications, while neural network alternatives require extensive hyperparameter tuning or work only in toy datasets under limited settings. In this paper, we introduce HyperFast, a meta-trained hypernetwork designed for instant classification of tabular data in a single forward pass. HyperFast generates a task-specific neural network tailored to an unseen dataset that can be directly used for classification inference, removing the need for training a model. We report extensive experiments with OpenML and genomic data, comparing HyperFast to competing tabular data neural networks, traditional ML methods, AutoML systems, and boosting machines. HyperFast shows highly competitive results, while being significantly faster. Additionally, our approach demonstrates robust adaptability across a variety of classification tasks with little to no fine-tuning, positioning HyperFast as a strong solution for numerous applications and rapid model deployment. HyperFast introduces a promising paradigm for fast classification, with the potential to substantially decrease the computational burden of deep learning. Our code, which offers a scikit-learn-like interface, along with the trained HyperFast model, can be found at \url{https://github.com/AI-sandbox/HyperFast}.
\end{abstract}

\section{Introduction}
\label{introduction}
Many different machine learning (ML) methods have been proposed for the task of supervised classification \cite{duda2000pattern}, following a traditional two-stage methodology. The initial stage involves the optimization of a model using the training portion of a dataset. Several tuning iterations are performed with the aim of finding the hyperparameter configuration of the model that yields the best performance on the specific task. In the second stage, the model with the chosen hyperparameter setup is used for evaluation and inference on the test set. 
Training and tuning models for classification tasks is time-consuming, and it often requires extensive data pre-processing, expertise in selecting hyperparameters that could fit the task at hand, and a validation process. 
Further, the computational and temporal costs of the traditional process can be prohibitive, particularly in real-time and large-scale applications, such as healthcare \cite{esteva2019guide}, or applications where rapid model deployment is necessary~\cite{deiana2022applications}, including data streaming, where models need to be updated or re-trained frequently. 
In this work, we propose HyperFast, a novel method to solve classification tasks from multiple domains with a single forward pass of a hypernetwork. We substitute the slow training stage of a classification network with a fixed hypernetwork 
that has been pre-trained (meta-trained) to predict the weights of a smaller neural network (i.e. main network) that can instantly solve the classification task with state-of-the-art performance. 
Recently, TabPFN \cite{hollmann2023tabpfn} has been proposed, introducing a pre-trained Transformer that is able to perform classification without training. However, it is constrained to $\leq$ 1000 training examples, 100 features and 10 classes, which limits its application to most real-world scenarios.
In this study, we are particularly interested in ensuring adaptability to large dataset sizes, 
filling the gap present in the current landscape of pre-trained models for instant tabular data classification. Our model is designed to work with both large and small datasets, while also providing adaptability to different numbers of samples, features, and categories. 

During the meta-training stage, the hypernetwork parameters are learnt and the parameters of a main model are inferred, that is, we are ``learning to learn'' from a wide variety of datasets (meta-training datasets) from different modalities for which HyperFast generates a smaller neural model that performs the actual classification. 
During the meta-testing or inference stage, HyperFast receives a ``support set'' of an unseen dataset (both features and labels), and predicts a set of weights for the main model, which classifies the test samples of the dataset.
In this way, the process of adapting the model to a new dataset is accelerated, and the model that does the meta-learning is decoupled from the model that does the actual inference on the data. In other words, we train a high-capacity meta-model to encode task-specific characteristics in the weights of a smaller model. 
Model size is also decoupled, which means that a large meta-learner can be trained just once, while many lightweight models generated by the meta-learner can be used for deployment in different applications such as edge computing, IoT devices, and mobile devices, where computational resources are constrained, and fast inference is indispensable. 
These properties are also helpful to accelerate production, improve privacy aspects, or for federated learning \cite{yang2019federated}. 
The meta-learner can instantly generate a model that is ready for deployment, but the generated weights might not be optimal. Thus, we also explore further improvements to quickly boost the performance before deployment and leverage all the power of the framework. For example, ensembles of multiple generated models can be used, or the generated weights can be used as an initial point for fine-tuning. More detail on many of the possibilities to improve model performance and obtain a stronger predictor can be found in the Appendix. 

The hypernetwork is trained on a wide range of datasets with different data distributions, allowing it to learn relevant and general meta-features, such that during testing the hypernetwork can adapt and predict an accurate set of weights for new unseen datasets. 
We evaluate the performance of HyperFast across a set of 15 tabular datasets, including genomics datasets and a standardized suite of datasets from OpenML \cite{bischl2021openml}. We also analyze the performance of HyperFast on larger time budgets by ensembling main networks generated with multiple forward passes and fine-tuning on inference. We compare our model to similar approaches and classical methods, both in terms of performance and time. Our method achieves competitive results compared to standard ML and AutoML algorithms tuned for up to one hour for each test dataset.

\section{Related Work}
\label{sec:relatedwork}

\paragraph{Hypernetworks.}
Building from evolutionary algorithms, HyperNEAT \cite{stanley2009hypercube} evolves Compositional Pattern-Producing Networks (CPPNs) to augment the weight structure for a larger main network. Based on this idea, \cite{ha2017hypernetworks} propose hypernetworks, where one neural network is used to generate weights for another neural network. The hypernetwork is trained end-to-end jointly with the main network to solve the task, producing weights in a deterministic way. 
\cite{krueger2018bayesian} and \cite{louizos2017multiplicative} propose variational approximations for weight generation using normalizing flows, \cite{deutsch2018generating} use multilayer perceptrons (MLPs) and convolutions, and \cite{ratzlaff2019hypergan} use generative adversarial networks (GANs). \cite{schurholt2022hyperrepresentations} explores unsupervised weight generation through model datasets. 
\cite{ashkenazi2022nern} use neural representations similar to NeRF \cite{mildenhall2020nerf} to reconstruct weights of a pre-trained network leveraging knowledge distillation. 
The HyperTransformer \cite{zhmoginov2022hypertransformer} is a few-shot learning hypernetwork based on the Transformer architecture that generates weights of a convolutional neural network (CNN). Unlike our method, the HyperTransformer is only designed for image classification and also requires training image and activation feature extractors.
HyperFast presents a novel approach by introducing hypernetworks for instant tabular classification. 
HyperFast solves the classification task by taking a set of labeled datapoints (support set) and generating the weights of a neural model that can be directly used to classify new unseen datapoints.
Previous work \cite{gidaris2018dynamic, qiao2018few} considered generating weights for specific layers (e.g., the last classification layer), while training the rest of the feature extractor. Here, we go one step further and consider generating all the weights of the model that performs the classification in a single forward pass.
Our hypernetwork design includes initial transformation modules, retrieval-based components, and different pooling operations in a unique architecture, offering feature permutation invariance and providing scalability and adaptability to new datasets while ensuring efficiency and speed. 

\paragraph{Meta-learning.} 
In the context of rapidly adapting to new tasks using limited data, meta-learning methods have emerged as powerful techniques. These approaches ``learn to learn'' by quickly integrating information at test time to make predictions for new, unseen tasks. A model $P_\theta(y|x,\mathcal{S})$ is learned for every new task, where $y$ is the target, $x$ is the test input, and $\mathcal{S} = \{X,Y\}, $ is the support set.
Metric-based learning methods such as Matching Networks \cite{vinyals2016matching} and Prototypical Networks \cite{snell2017prototypical} map a labelled support set $\mathcal{S}$ into an embedding space, where a distance is computed with the embedding of an unlabelled query sample to map it to its label. As in kernel-based methods, the model $P_\theta$ can be obtained through $P_\theta (y|x,\mathcal{S}) = \sum_{x_i,y_i\in \mathcal{S}} K_\theta (x,x_i)y_i $.
Optimization-based methods such as Model-agnostic meta-learning (MAML) \cite{finn2017model} learn an initial set of model parameters and perform an additional optimization through a function $f_{\theta(\mathcal{S})}$, where model weights $\theta$ are adjusted with one or more gradient updates given the support set of the task $\mathcal{S}$, i.e., $P_\theta (y|x,\mathcal{S}) = f_{\theta(\mathcal{S})} (x,\mathcal{S})$. Finally, model-based approaches such as Neural Processes (NPs) \cite{garnelo2018neural, garnelo2018conditional} first process both support samples and query samples independently as in Deep Sets \cite{zaheer2017deep}, and the predicted embeddings are aggregated with a permutation-invariant pooling operation, resulting in a dataset-level summary that is fed to a second stage network that predicts the output for the query sample. The overall model is defined by a function $f$ and the process can be mathematically described as $P_\theta (y|x,\mathcal{S}) = f_{\theta} (x,\mathcal{S})$. Similarly, TabPFN~\cite{hollmann2023tabpfn} learns to learn Bayesian inference by using a Transformer network.
In contrast, our method directly obtains the model weights $\theta$ in a single forward step through an independent network, i.e., the hypernetwork $h$, such that $P_\theta (y|x,\mathcal{S}) = f_{h(\mathcal{S})}(x)$.

\paragraph{Deep Learning for Tabular Data.}
Although deep learning (DL) models achieve state-of-the-art results in many domains (e.g., language, computer vision, audio), this is not the case for tabular data. Tree-based models such as XGBoost \cite{xgboost}, LightGBM \cite{lightgbm} or CatBoost \cite{catboost} are still the preferred choice in some tabular data applications \cite{grinsztajn2022why, shwartz2022tabular}. AutoML methods \cite{he2021automl, feurer-arxiv20a, erickson2020autogluon} are also a popular alternative, automatically selecting the most appropriate ML algorithm and its hyperparameter configuration. However, it has been shown that there is not a universal superior solution \cite{gorishniy2021revisiting, mcelfresh2023neural}, and many deep learning approaches for tabular data have been proposed \cite{kadra2021cocktails, arik2021tabnet, somepalli2021saint, kossen2021self, gorishniy2021revisiting, yan2023t2g, chen2022tabcaps, katzir2020net, popov2019neural, hollmann2023tabpfn, chen2022danets, chen2023excelformer, zhu2023xtab, zhang2023generative}. 
\cite{kadra2021cocktails} introduced Regularization Cocktails, where different regularization techniques are applied to simple MLPs to boost performance. 
Recent work has explored using attention mechanisms to improve performance on tabular data. TabNet \cite{arik2021tabnet} adopts sequential attention on subsets of features, SAINT \cite{somepalli2021saint} applies attention over rows and columns in a BERT-style fashion and uses contrastive pre-training with data augmentation, 
NPT \cite{kossen2021self} introduces attention between data points, ExcelFormer~\cite{chen2023excelformer} models feature interaction and feature representation alternately, FT-Transformer~\cite{gorishniy2021revisiting} adapts a Transformer with embeddings for categorical and numerical features, and T2G-Former~\cite{yan2023t2g} includes a graph estimator to guide tabular feature interaction. TabCaps explore capsule networks~\cite{chen2022tabcaps}, Net-DNF~\cite{katzir2020net} disjunctive normal formulas, NODE~\cite{popov2019neural} combines ensembles of differential oblivious decision trees with multi-layer hierarchical representations, DANets~\cite{chen2022danets} learn groups of correlative input features to generate higher-level features, and other works explore large language models (LLM) for tabular data pre-training~\cite{zhu2023xtab, zhang2023generative}. 
Nevertheless, most of the proposed DL models for tabular data require slow training and custom hyperparameter tuning for every new dataset. In contrast, we focus on off-the-shelf models that do not need extensive tuning for a new task. 
In this direction, TabPFN \cite{hollmann2023tabpfn} pre-trains a Transformer on synthetic data given a prior to perform tabular data classification in a single forward pass with no hyperparameter tuning. 
However, TabPFN can only be applied to small tabular datasets, i.e., $\leq$ 1000 training examples, 100 features and 10 classes. 

\section{Background}

\subsection{Meta-Learning Problem Setting}

In our meta-learning experiments, we train a model $h$ (i.e., the hypernetwork) that is able to quickly adapt to new tasks given some observations, and generate the weights of a main model $f$ that solves the task for unseen datapoints. We consider a set of classification tasks $\mathcal{T}$ where each task $t\in\mathcal{T}$ is associated with a \emph{support set} $\mathcal{S}_t$ of examples that are sufficient to find the optimal model $f$ that solves the task, a loss function $\mathcal{L}_t$, and a \emph{query set} $\mathcal{Q}_t$ to define $\mathcal{L}_t$. 
The first phase is the \emph{meta-training}, where in each step a different training task $t\in\mathcal{T}_{\text{meta-train}}$ is selected. 
We compile a set of meta-training datasets $\mathcal{D}_{\text{meta-train}}$, where each dataset $d\in\mathcal{D}_{\text{meta-train}}$ is composed of a training set $d_{\text{train}}$ and a test set $d_{\text{test}}$, as in the common machine learning setup. In each meta-training step, a task $t\in\mathcal{T}_{\text{meta-train}}$ is sampled by first randomly choosing a meta-training dataset $d$. Then, $\mathcal{S}_t$ and $\mathcal{Q}_t$ are generated by sampling examples from $d_{\text{train}}$ and $d_{\text{test}}$, respectively.
Meta-validation is also performed intermittently through meta-training, where a separate set of meta-validation datasets $\mathcal{D}_{\text{meta-val}}$ is used to generate validation tasks $\mathcal{T}_{\text{meta-val}}$ to evaluate our algorithm and select the best performing model.

Once HyperFast is trained, an independent set of meta-testing datasets $\mathcal{D}_{\text{meta-test}}$ are used to create the evaluation tasks $\mathcal{T}_{\text{meta-test}}$ in which the selected model is evaluated. This approach allows us to extend the classical ``$n$-way-$k$-shot'' few-shot learning setting to handle multiple datasets with varying distributions and categories, testing the robustness and generalization of our model on new data.

As opposed to the training tasks, where each $t\in \mathcal{T}_{\text{meta-train}}$ is randomly generated at every meta-training step, $\mathcal{T}_{\text{meta-val}}$ and  $\mathcal{T}_{\text{meta-test}}$ are sets of partially fixed tasks, as the query set $\mathcal{Q}_t$ always covers all $d_{\text{test}}$ samples, in order to evaluate and compare with other methods equally, which also tests their performance on the entire test subset $d_{\text{test}}$ of a dataset $d$.

\begin{figure*}[!t]
    \centering
    \includegraphics[width=0.975\textwidth]{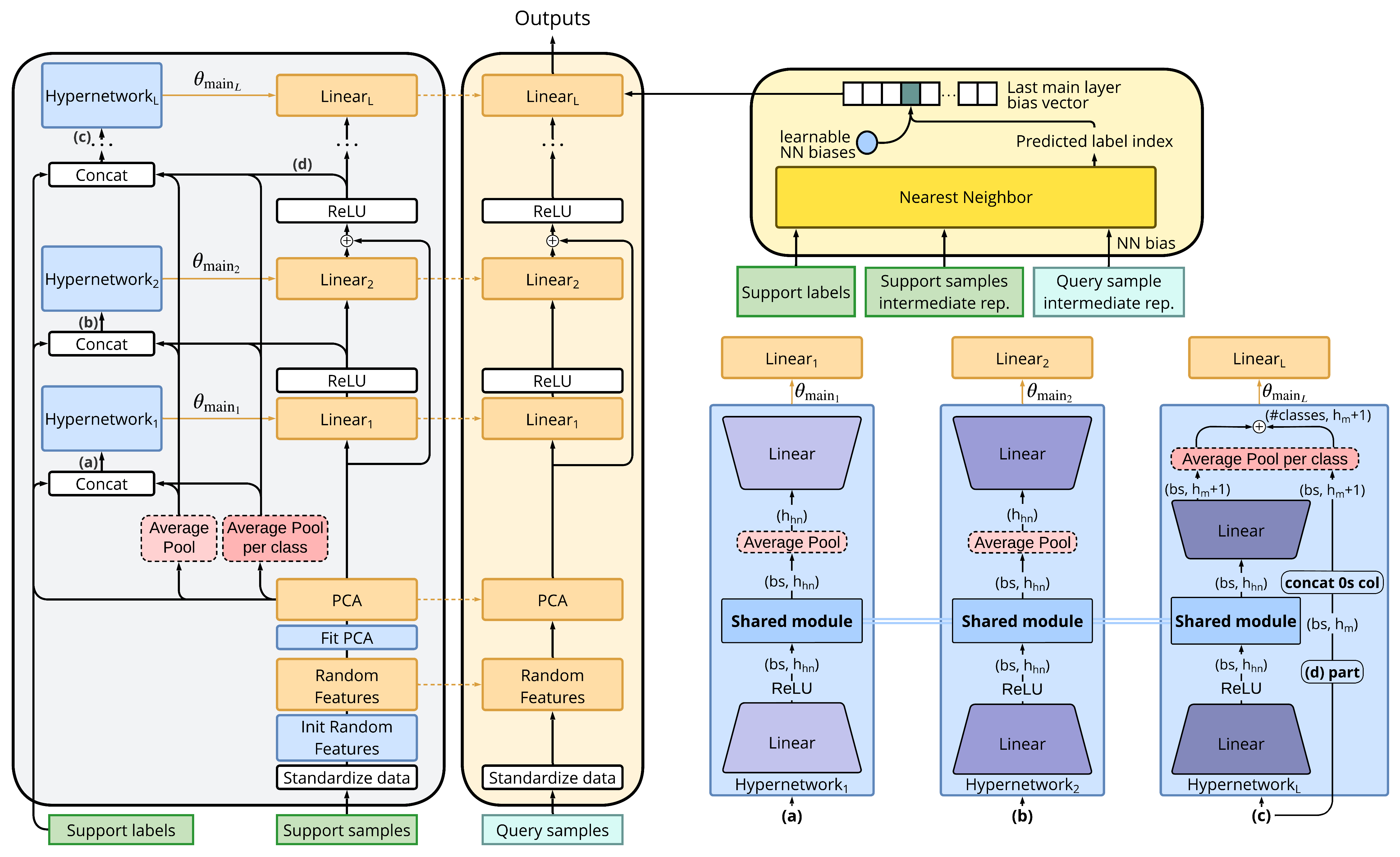}
    \caption{(left) HyperFast framework. (right) Architecture detail. Each hypernetwork module receives representations of the support set of \emph{batch size} ($bs$) samples. The modules $l \in [1, L-1]$ compress the representations into a single embedding of \emph{hypernetwork hidden size} ($h_{\text{hn}}$) to then generate the main network weights $\theta_{\text{main}_l}$. Module $L$ summarizes the representations per class with embeddings of \emph{main hidden size} $(h_{\text{m}})+1$, directly obtaining the weights of the last classification layer $\theta_{\text{main}_L}$.}
    \label{figure1}
\end{figure*}

\section{HyperFast}
\label{hyperfast}

The traditional training process can be seen as a function $f(X,Y)=\theta$, that receives training instances $X \in \mathbb{R}^{N\times D}$ and corresponding labels $Y\in \mathbb{R}^{N}$, and produces a set of trained weights $\theta$ of a model. 
In this work, we substitute the training process with HyperFast, a pre-trained meta-model based on a hypernetwork \cite{ha2017hypernetworks} $h$, that takes as input a subset of the training data (i.e. support set $\mathcal{S}_t$) for a task $t\in\mathcal{T}_{\text{meta-train}}$ and predicts the weights of a main neural network $f_\theta$ for the given task $t$ as $\theta^*=h(\mathcal{S}_t)$.
The target model $f_{h(\mathcal{S}_t)}$ directly uses the predicted weights and makes predictions for test data points $x \in \mathcal{Q}_t$ in a single forward pass, such that $P_\theta (y|x,\mathcal{S}) = f_{h(\mathcal{S})}(x)$.

The meta-model is learnt by observing a set of tasks $t\in\mathcal{T}_{\text{meta-train}}$ and minimizing $\mathcal{L}_t(f_{h(\mathcal{S})}(x))$. In this section, we detail the design and architecture of $h$, named HyperFast in analogy to Hypernetworks \cite{ha2017hypernetworks}, and the ability to instantly adapt to new datasets in a single forward pass. \cref{figure1} illustrates the HyperFast framework and the main building blocks of the architecture. HyperFast is a multi-stage model with initial transformation layers that allows variable input size and permutation invariance, and a combination of linear layers and pooling operations that take both support samples and their associated labels to directly predict the weights $\theta_{\text{main}_l}$ (weight matrix and bias) of linear layers $l\in[1,L]$ of a target neural network. 
All trainable modules of HyperFast are learnt end-to-end by optimizing the classification loss of the main network evaluated on $\mathcal{Q}_t$.

The framework and HyperFast architecture proposed in this paper is a specific instance of a more general framework that could be easily extended to predict convolutional layers, batch normalization layers, recurrent layers, or deeper networks, for example. However, the architecture design depicted in \cref{figure1}
selection has been driven by a global and simple approach to handle a wide range of multi-domain data, while seeking efficiency and speed.

\subsection{Initial Transformation Layers}
\label{ssec:rfpca}
 
Properly dealing with datasets of differing dimensionality is a challenge, and one common solution is to apply padding~\cite{hollmann2023tabpfn} or to keep a subset of selected features up to a fixed size. 
We first perform a general data standardization stage by one-hot encoding categorical features, mean imputing missing numerical features, mode imputing missing categorical features, and feature-wise transforming to zero mean and unit variance. Then, HyperFast comprises initial layers that project datasets of different dimensionality to fixed-size and feature-permutation invariant representations. The kernel trick can be used to project data to a Reproducing Kernel Hilbert Space (RKHS) \cite{aronszajn1950theory} when the number of dimensions tends to infinity. 
However, this would require computing all pair-wise kernel distance in every step of the training process. Instead, we use random features (RF) \cite{rahimi2007random} to approximate a kernel with a fixed and finite number of dimensions. Random features are computed as $\phi(x)=a(Wx)$, where $a(\cdot)$ is a non-linearity, and $W$ is a random projection matrix that follows a pre-defined distribution. 
The approximated kernel depends on the distribution of $W$ and the selected non-linearity. In our case, we sample $W$ from a Gaussian distribution and use the ReLU activation as non-linearity, approximating an arc-cosine kernel. We choose to approximate the arc-cosine kernel because it captures sparse, neural network-like feature representations in a non-parametric kernel setting~\cite{cho2009kernel}. In contrast, polynomial kernel's features are neither sparse nor non-negative, and radial basis function (RBF) kernels capture localized similarities. 
In each forward step, the random features projection matrix is re-initialized and sampled. The number of rows is adjusted to match the dimensionality of the input dataset, while the number of columns remains fixed, determining the output size.

The combination of random features with Principal Component Analysis (PCA) provides an efficient low-rank randomized approximation of Kernel PCA~\cite{sriperumbudur2017approximate, lopez2014randomized}. 
We estimate the PCA parameters $\psi$ using the support set and project the data onto a specified number of components. Subsequently, both $\phi$ and $\psi$ are applied to the query samples to transform the data. This transformed data is then forwarded through the $L$ generated linear layers of the main network.

\subsection{Hypernetwork Modules}
\label{hnarch}

The process of generating the weights of the main network is done layer-by-layer, by multiple hypernetwork modules with both shared and layer-specific parameters, see \cref{figure1}. The hypernetwork module that generates the weights for the main network layer $l$ receives as input the representations of the support samples in the previous stage, concatenated with the one-hot encoded support labels, the global average, and the class average of the low-rank Kernel PCA projection of the support set. Note that each sample is concatenated with the class average corresponding to its associated label. 

For predicting $\theta_{\text{main}_1}$, the representations are the low-rank Kernel PCA projection of the support samples. For $\theta_{\text{main}_l} \in [2, L]$, the hypernetwork module receives the intermediate representations of the support set in the main network at the output of layer $l-1$, after non-linearities and residual connections are applied. \cref{figure1} represents a specific multi-layer perceptron (MLP) architecture with ReLU activations and residual connections, which we use for our experiments. However, the HyperFast framework can be easily extended to generate weights for other main network architectures.

The hypernetwork modules that predict the layers $l\in[1, L-1]$ are composed of MLPs with shared middle layers that take the support set representations and labels, and output embeddings for each sample in the support set. Then, permutation-invariant weights are obtained averaging all support embeddings in a similar fashion to Deep Sets \cite{zaheer2017deep}, to obtain a single dataset embedding that is passed to a final linear layer which outputs the final weights $\theta_{\text{main}_l}$ of $l$. $\theta_{\text{main}_l}$ is then reshaped as weight matrix and bias vector to forward the data through the main network.

Layer $L$ is the classification layer of the main layer that outputs the logits for the final prediction. In this case, the intermediate representations after the layer $L-1$ and labels information are encoded through a MLP hypernetwork but the weights $\theta_{\text{main}_L}$ are not directly predicted from a global embedding. Instead, we leverage the fact that the rows of the classification layer weight matrix correspond to the different categories of the task. We perform an average pooling per class, and obtain the rows of the classification weight matrix (and bias) as the average of representations for each category. 
This also allows a much lightweight implementation, instead of directly predict the weight matrix. Additionally, we add a residual connection~\cite{he2016deep} from the previous layer representations for which we also perform a per class average, which helps in retaining category information from the input.
Finally, we consider a module based on Nearest Neighbors to add learnable parameters (NN biases) to the classification layer bias vector of the main network. The label of a query sample is predicted with NN using the support set and the intermediate representations of the data across the main network, such that $P_\theta (y|x,\mathcal{S}) = f_{h(\mathcal{S})}(x,\mathcal{S})$. We consider the representations after the PCA projection, and after each linear layer. The NN biases are added to the position of the bias of the last main classification layer corresponding to the predicted label.

Once the main network is fully generated, query samples can be forwarded to make predictions. During meta-training, the predictions for the query samples $\mathcal{Q}_t$ of $t \in \mathcal{T}_{\text{meta-train}}$ are used to compute the cross-entropy loss $\mathcal{L}_t$ and learn the parameters of HyperFast end-to-end. In evaluation, all hypernetwork parameters are frozen and generate weights for a main network in a single forward pass.

\section{Experiments}
\label{experiments}

In this section, we compare HyperFast to many standard ML methods, AutoML systems and DL methods for tabular data on a wide variety of tabular classification tasks, listed in the Appendix. We do not perform any hyperparameter tuning to HyperFast, as it can be used as an off-the-shelf hypernetwork ready to generate networks to perform inference on new datasets. We then compare the performance and runtime of the generated model in a single forward pass, as well as the combination of multiple generated networks by increasing the ensemble size and fine-tuning on inference.

 \begin{figure*}[ht]
  \centering
  \includegraphics[width=\textwidth]{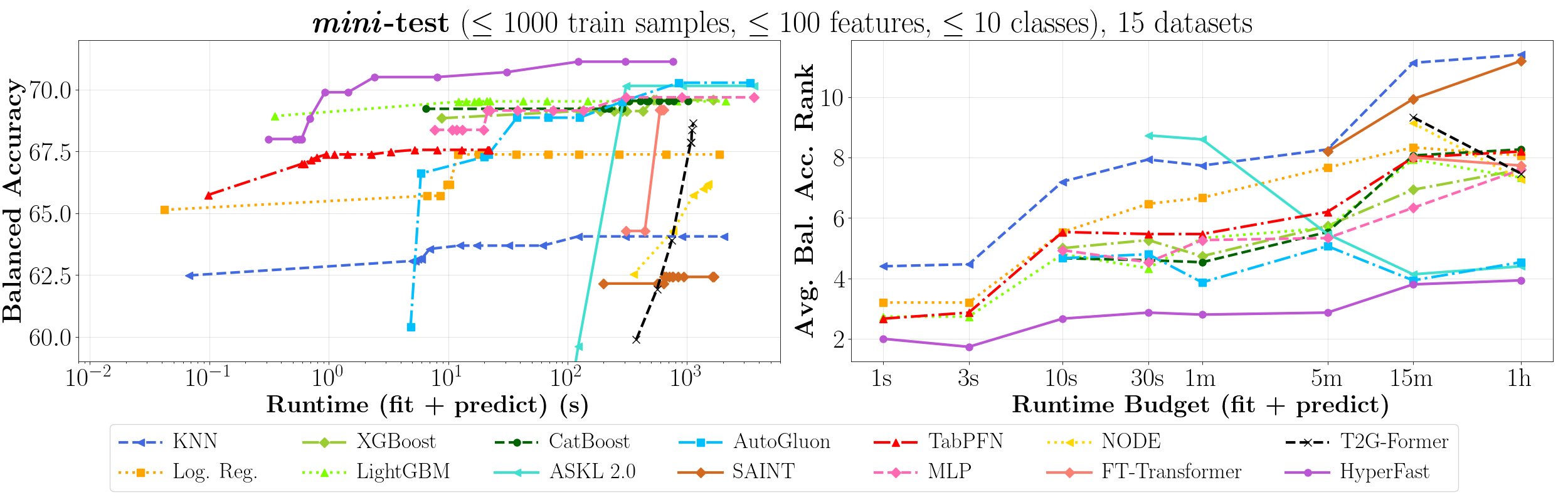}
  \caption{Runtime (fit + predict) vs. performance and average rank for given runtime budgets on the \emph{mini}-test (small-sized version of the 15 meta-test datasets with $\leq$ 1000 training examples, $\leq$ 100 features and $\leq$ 10 classes restrictions).}
  \label{fig:mini_test_plots}
\end{figure*}

\paragraph{Baselines}
We compare HyperFast to standard ML methods, AutoML systems and state-of-the-art DL methods for tabular data. We first consider simple and fast ML methods as $K$-Nearest Neighbors (KNN) and Logistic Regression (Log. Reg.), and a MLP matching the architecture of the target network. We also evaluate against tree-based boosting methods: XGBoost \cite{xgboost}, LightGBM \cite{lightgbm}, and CatBoost \cite{catboost}. As AutoML methods we incorporate Auto-Sklearn 2.0 (ASKL 2.0)~\cite{feurer-arxiv20a}, which uses Bayesian Optimization to efficiently discover a top-performing ML model or a combination of models by ensembling, and AutoGluon \cite{erickson2020autogluon}, which uses a selection of models such as neural networks, KNN, and tree-based models, combining them into a stacked ensemble. Finally, we include popular tabular DL methods: SAINT \cite{somepalli2021saint}, TabPFN~\cite{hollmann2023tabpfn}, NODE~\cite{popov2019neural}, FT-Transformer~\cite{gorishniy2021revisiting}, and T2G-Former~\cite{yan2023t2g}. 
All standard ML models, gradient boosting methods and SAINT are evaluated using 5-fold cross validation for hyperparameter adjustment. Hyperparameter configurations are drawn from search spaces (detailed in the Appendix) unitl 10\,000 configurations are explored, a specified time budget is reached, or more than 32 GB of memory are required if GPU training is possible for the model. Then, the model is trained on the full training set with the best configuration between the hyperparameter search result and the default. For the AutoML methods, the time budget is given. Finally, both TabPFN and our HyperFast are pre-trained models with no hyperparameter tuning requirements, but with ensembling capabilities. Thus, we perform ensembling for each method until a given number of members are used (detailed in the Appendix) or until 32 GB of GPU memory are overloaded.

\paragraph{Data}
\label{data}

We collect a wide variety of datasets from different modalities. We use the 70 tabular datasets from the OpenML-CC18 suite~\cite{bischl2021openml} which, to the best of our knowledge, is the \emph{largest} and most used standardized tabular dataset benchmark, composed of standard classification datasets (e.g., Breast Cancer, Bank Marketing). 
The collection of OpenML datasets is randomly shuffled and divided into meta-training, meta-validation and meta-testing sets, with a 75\%-10\%-15\% split, respectively. 
We also include tabular genomics datasets sourced from distinct biobanks. Specifically, we utilize genome sequences of dogs \cite{bartusiak2022predicting} for dog clade (group of breeds) prediction in meta-training, European (British) humans from the UK Biobank (UKB) \cite{sudlow2015uk} for phenotype prediction in meta-validation, and HapMap3~\cite{international2010integrating} for subpopulation prediction in the meta-test. This strict separation ensures we meta-learn and evaluate on substantially different distributions and tasks. More details on the processing of these datasets are provided in the Appendix. 
The simple ML methods, implemented with scikit-learn \cite{scikit-learn}, and the MLP, receive the numerical features standardized with zero mean and unit variance, and the categorical features are one-hot encoded. For the missing values, we perform mean imputation for numerical features and mode imputation for categorical features, as it was the configuration that yielded the best performance. We also perform imputation of missing values for SAINT, NODE, and FT-Transformer.
Boosting methods, AutoML systems, and TabPFN receive the raw data and the indices of categorical features when needed, as their documentation states that they pre-process inputs internally. 

Apart from the large-sized original test datasets, we create a secondary small-sized tabular data version (\emph{mini}-test) of the meta-testing datasets to compare to TabPFN, as it is only able to handle $\leq$ 1000 training examples, $\leq$ 100 features and $\leq$ 10 classes. We randomly select a subset of $\leq$ 1000 training samples and $\leq$ 100 features for each dataset. 
We do not perform any downsizing in terms of number of classes as the highest number of classes appearing in the meta-testing set is 10. However, HyperFast is pre-trained with datasets with higher number of classes and can be used in inference for datasets with $>$10 classes. 
Only models that can complete the runs for all 15 datasets in less than 48 hours in their default configuration are included in our large-scale experiments. 
The experiments, which are conducted for all models and both size versions of the 15 meta-testing datasets, considering all time budgets shown in \cref{fig:mini_test_plots}, require a total of 2 months to complete. Therefore, we show additional results with 10 repetitions of the experiments for a specific time budget of 5 minutes for each dataset in the Appendix.

\paragraph{Experimental setup}
\label{setup}

We perform supervised classification with HyperFast and all other baselines on the \emph{mini}-test, a small-sized version of the meta-test datasets $\mathcal{D}_{\text{meta-test}}$, and in the original large-scale datasets. To train HyperFast, we use a different set of meta-training datasets, $\mathcal{D}_{\text{meta-train}}$, and select the model with the best average performance on the meta-validation datasets, $\mathcal{D}_{\text{meta-val}}$. We report balanced accuracy, which is the mean of sensitivity and specificity. Balanced accuracy provides a more objective and robust evaluation across classes, especially in the context of imbalanced datasets. In contrast, standard accuracy can be misleading, often masking poor performance in minority classes. 
We evaluate the models on a time budget (including tuning, training, and prediction) to correctly assess computational complexity and performance. 
The average rank is also reported. 

In order to transform the data to a fixed-size and permutation invariant representation, we apply Random Features and Principal Component Analysis to both support samples and query samples. We set a Random Features projection to 32\,768 ($2^{15}$) features, sampled from a normal distribution following the He initialization \cite{he2015delving}, followed by a ReLU activation. Note that the random linear layer that computes the random features is not trained, and re-initialized in each HyperFast forward step.
Then, we keep the principal components (PCs) associated to the 784 largest eigenvalues, as many of the datasets considered have this dimensionality, and it is a more than sufficient number of dimensions to retain the important information of higher dimensional datasets while preserving efficiency.
After the PCA projection, most genomics datasets resemble a similar histogram distribution (i.e., zero mean, small deviation and no outliers). However, it is not the case for some OpenML datasets, which are also centered around zero but present many outliers. Thus, we clip the data after PCA at $4\sigma$. 

 \begin{figure*}[ht]
  \centering
  \includegraphics[width=\textwidth]{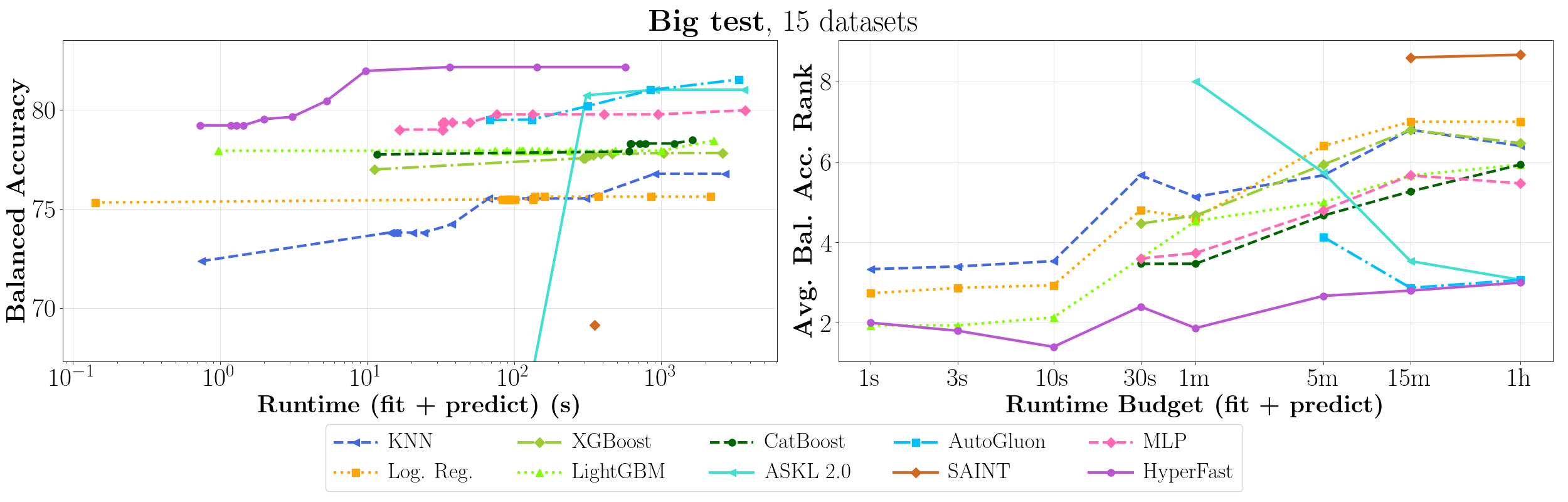} 
  \caption{Runtime (fit + predict) vs. performance and average rank for given runtime budgets on the big test: 15 large/medium-sized meta-datasets.}
  \label{fig:big_test_plots}
\end{figure*}

The hypernetwork modules receive a concatenation of intermediate representations of the support samples, and the support labels. Given that each dataset features a different number of categories and linear layers require a fixed input size, we one-hot encode the labels and apply zero padding up to the maximum number of categories considered in the experiments. It is important to note that the number of categories that HyperFast can handle is easily extendable by expanding the input size of HyperFast and zero padding the remaining input dimensions. Such modifications have a negligible impact on efficiency or memory requirements, up to a reasonable number of categories. 
As a shared module we use 2 feed-forward layers with a hidden size of 1024 and ReLU activations. 
For the main network, we consider a 3-layer MLP with a residual connection~\cite{he2016deep}, and a main network hidden size equal to the number of PCs (784 dimensions). 
We select this simple architecture to be able to obtain competitive performance on a wide variety of datasets with a single trained model while preserving efficiency. Other alternatives include predicting weights for CNN layers for only-image datasets, or recurrent layers, for sequential data. Instead, we create a general and simple meta-learning framework to perform fast lightweight inference.
In the NN bias module, we randomly select a subset of a maximum of 2048 support samples, since computing all pairwise distances for a large number of datapoints causes high inefficiency and GPU memory overload.
A maximum batch size of 2048 samples is used for training, and we make sure to have a sufficient number of samples per category in every case.

 In evaluation, we show prediction results with HyperFast in a single forward pass, as well as predictions by ensembling main networks generated in multiple forward passes. We also experiment with performing gradient steps with the training data of the meta-testing datasets on the generated main networks, including the random features projection matrix, PCA parameters and linear layers.

\paragraph{Results on small-sized datasets.}
\label{results_small}
We first compare HyperFast to the other methods on a small-sized setting with datasets having $\leq$ 1000 training samples, $\leq$ 100 features and $\leq$ 10 classes, in order to compare to TabPFN. As shown in \cref{fig:mini_test_plots}, HyperFast delivers superior results in both performance and runtime, with better prediction capabilities up to 3 orders of magnitude faster than competing methods. Simple ML methods such as KNN and Log. Reg. also deliver instant predictions, but do not achieve remarkable performance. 
Interestingly, an MLP (with an architecture identical to the network generated by HyperFast, including the including the initial transformation layers) performs on par with XGBoost. 
However, HyperFast surpasses gradient-boosting techniques in both runtime and performance.  LightGBM stands out as the only boosting machine that achieves a higher balanced accuracy in a similar runtime to a single forward pass by HyperFast. 
Yet, an ensemble of networks generated by HyperFast outperforms all fine-tuned boosting machines in under 3 seconds. 
TabPFN is noted for its rapid predictions and outperforms NODE and SAINT. But on average, it falls behind gradient boosting machines and neural models, including HyperFast. FT-Transformer, T2G-Former, NODE, and SAINT are DL tabular models with very time-consuming training, and FT-Transformer obtains the highest performance among them, similar to that of gradient-boosting machines. 
AutoML systems are superior to the other baselines when given higher runtime budgets. However, HyperFast still outperforms both AutoGluon and ASKL 2.0 for runtimes up to 1h, obtaining the lowest rank throughout all the budgets in the mini test.

\begin{table*}[!ht]
  \centering
\begin{tabular}{lrrrrrr}
\toprule
                          Variation &  Bal. acc. (\%) &  Bal. acc. diff. &  Fit time (s) &  Pred. time (s) & HF size & Model size\\
\midrule
                      Base model (784 PCs) &          81.496 &            - &         0.600 &               0.125  & 1.27 B & 52.65 M\\ 
                             No RF         &          75.387 &           -6.108 &         0.126 &           0.114  & 1.27 B &  1.85 M\\
                          No RF-PCA        &          73.704 &           -7.792 &         0.029 &           0.109  & 1.26 B &  1.23 M\\ 
                        First 512 PCs only &          81.347 &           -0.149 &         0.625 &           0.125  & 547 M & 43.03 M\\
                        First 256 PCs only &          81.235 &           -0.261 &         0.640 &           0.042  & 140 M & 34.25 M\\
        $d_{\text{RF}}$=16\,384 ($2^{14}$) &          81.059 &           -0.436 &         0.510 &           0.116  & 1.27 B & 26.95 M\\
          No concat PCA to hypern. modules &          80.727 &           -0.769 &         0.583 &           0.125  & 1.26 B & 52.65 M\\
           1 linear layer in shared module &          80.790 &           -0.706 &         0.637 &           0.125  & 1.27 B & 52.65 M\\
          No residual conn. in hypern.$_L$ &          77.835 &           -3.660 &         0.620 &           0.125  & 1.27 B & 52.65 M\\
            No residual con. in main model &          80.318 &           -1.178 &         0.633 &           0.125  & 1.27 B & 52.65 M\\
             No NN bias using PCA features &          81.305 &           -0.191 &         0.625 &           0.125  & 1.27 B & 52.65 M\\
             No NN bias using interm. act. &          80.703 &           -0.793 &         0.628 &           0.125  & 1.27 B & 52.65 M\\
                              No NN biases &          79.714 &           -1.781 &         0.628 &           0.125  & 1.27 B & 52.65 M\\
           Random init. linear layers main &          72.229 &           -9.267 &         0.437 &           0.125  & -      & 52.65 M\\
\bottomrule
\end{tabular}
  \caption{Ablation studies on HyperFast performing a single forward pass. Time results are shown for a single GPU. \emph{HF size} denotes the number of trainable parameters of HyperFast, i.e., the meta-model, while \emph{Model size} denotes the size of the generated model.}
  \label{tab:ablation}
\end{table*}

\paragraph{Results on medium/large-scale datasets.}
\label{results_large}
\cref{fig:big_test_plots} benchmarks the algorithms on large real-world datasets. We observe that HyperFast is able to obtain predictions in less than a second, and achieves the overall best performance in a wide range of runtime budgets, ranging from 1 second to 5 minutes. For more extended budgets, up to 1h per dataset, HyperFast's performance is on par with other AutoML systems. 
Specifically, HyperFast, ASKL 2.0, and AutoGluon all achieve an average rank of approximately 3.0. 
In comparison, gradient-boosting machines plateau at a balanced accuracy of 78.4\% and rank above 5.9, being outperformed by the MLP. SAINT obtains the lowest performance, using the hyperparameter configuration that the authors implement for the biggest datasets they consider in their benchmark. No hyperparameter optimization is performed for SAINT in the big test since larger architectures do not fit in GPU memory for the larger datasets. Additional experiments with very high-dimensional genomic datasets can be found in the appendix.

\paragraph{Ablation Experiments}
\label{sec:ablation}

In \cref{tab:ablation}, we present ablation studies for the HyperFast framework, exploring variations affecting both hypernetwork modules and the generated model. 
First, we consider removing the RF and both RF and PCA modules, obtaining a fixed-sized input by keeping the first 784 features or applying zero padding. The weight generation time is reduced from 0.6s to 0.12s and 0.03s, since the main time bottleneck is the RF matrix multiplication and SVD to obtain the PCs. Also, the main model size is greatly reduced as RFs account for most parameters, but the drop in performance is one of the most significant. This is because RF and PCA not only allow transforming any dataset to a fixed number of features, but also homogenize the input data to HyperFast and the generated network across datasets. For example, the first feature post RF-PCA holds the most variance, with subsequent features capturing the maximum variance that is orthogonal to the previous dimensions, with minimal information loss. Also, histogram distributions are similar across datasets with zero mean. These properties help in learning important meta-features across different dataset distributions. If we scale down RF-PCA by reducing the number of PCs used and the RF dimensionality, we observe that model size is significantly reduced while the drop in performance is not critical, which shows that most dataset relevant information is preserved, even using 512 or 256 PCs. These observations can help create even more efficient HyperFast desings in the future. 
In addition, PCA representations concatenated to hypernetwork inputs retain key information without a major parameter increase. We also observe that reducing the shared hypernetwork module from 2 to 1 layer degrades performance, and residual connections in both the hypernetwork and main model are key to retain post-PCA and per class information, while not increasing model size and runtime. 
We also analyze the retrieval-based component of HyperFast. We observe that NN biases in the last classification layer improve predictions while maintaining model size, especially using the intermediate activations of the main network as features. Finally, if we replace the weights produced by HyperFast by random weights, and base the prediction solely on the Nearest Neighbor-based component, we observe the biggest drop in performance.

\section{Conclusion}
\label{conclusion}

We present HyperFast, a meta-trained hypernetwork designed to perform rapid classification of tabular data by encoding task information in the prediction of the weights of a target network in a single forward pass. 
Our experiments show that HyperFast consistently improves performance over traditional ML methods and tabular-specific DL architectures in a matter of seconds. Remarkably, it is able to replace the traditional training of a neural network, and achieves competitive results with state-of-the-art AutoML frameworks trained for 1h.
HyperFast eliminates the necessity for time-consuming hyperparameter tuning, making it a highly accessible, off-the-shelf model that can be specially useful for fast classification tasks. We also explore how we can leverage all training data by creating ensembles of generated networks and fine-tuning them on inference, significantly boosting performance at almost no additional computational cost. 
Future work should consider expanding this framework to a general architecture or multi-hypernetwork setting that is able to handle regression tasks, multi-domain and high-dimensional non-tabular settings such as audio streams, 3D, and video.

\section{Acknowledgments}
This work was partially supported by a grant from the Stanford Institute for Human-Centered Artificial
Intelligence (HAI) and by NIH under award R01HG010140. This research has been conducted using
the UK Biobank Resource under Application Number 24983.

\bibliography{refs}

\clearpage
\appendix

\section{Experimental Setup}
\subsection{Datasets}
\label{appendix-datasets}

\paragraph{OpenML}
We integrate the OpenML Curated Classification benchmarking suite 2018 (OpenML-CC18) \cite{bischl2021openml}. OpenML-CC18 consists of 72 diverse and curated classification tasks, and we keep 70 datasets, excluding the vision datasets Fashion-MNIST and CIFAR-10. We split each dataset into 80\% train and 20\% test.

\paragraph{HapMap3}
HapMap3 \cite{international2010integrating} is a publicly available dataset that contains single-nucleotide polymorphisms (SNPs) sequences of whole-genome data from humans with subpopulation annotations. Samples are filtered for the 10 largest human subpopulations, which are used as categories for the \emph{hapmap} datasets. Individuals are split into 75\% for train and 25\% for test. SNPs with missing values for any sample are discarded. Finally, 5 different datasets are created by randomly sampling 784 SNP positions from different sections of the chromosomes, which are encoded as binary values. For every created dataset, labels are permuted to avoid overfitting to the positions of the labels of each subpopulation.

\paragraph{Dogs}
Similarly, Dogs \cite{bartusiak2022predicting} is a dataset of dog DNA sequences. The dataset consists of the genotyping array of purebred dogs from 75 breeds. Dog breeds can be organized into clades, which are groups of dog breeds that share a common ancestor. Since the number of samples per breed in the dataset is very low, breeds are clustered into clades, and the 10 most common clades are kept and used as categories for the \emph{dogs} datasets. Samples are split into 75\% for train and 25\% for test. SNPs with missing values for any sample are discarded. Finally, 30 different datasets are created by randomly sampling 784 positions from different sections of the chromosomes. For every created dataset, labels are permuted to avoid overfitting to the positions of the labels of each clade. 

\paragraph{UK Biobank}
The UK Biobank \cite{sudlow2015uk} is a large-scale biobank, from which we use the genotyping array data and full phenotypes as processed in \cite{qian2020fast}. We include 8 of the most predictive binary phenotypes according to their polygenic risk score (PRS) model: 

\begin{itemize}
    \item Hair colour (natural, before greying) red: \emph{red hair}
    \item Hair colour (natural, before greying) blonde: \emph{blonde hair}
    \item Hair colour (natural, before greying) dark brown: \emph{dark brown hair}
    \item Hair colour (natural, before greying) black: \emph{black hair}
    \item Ease of skin tanning (Never tan, only burn): \emph{skin burn}
    \item Ease of skin tanning (Get very tanned): \emph{skin tan}
    \item Hair colour (natural, before greying) brown: \emph{brown hair}
    \item Malabsorption/coeliac disease: \emph{malabsorption-coeliac}
\end{itemize}

In order to allow proper phenotype prediction modeling, it is a standard practice to stick to a single population, to avoid the prediction being biased by other factors.
In this case, we filter by the majority population in the UK Biobank, which is British individuals with European ancestry. Then, we create a balanced dataset for each phenotype by selecting all samples of the minority class (presence of the phenotype), and randomly selecting the same number of samples from the majority class. Variants (features) are selected based on the PRS model weights reported~\cite{tanigawa2022significant}. We split each dataset into 80\% train and 20\% test.

The collection of OpenML datasets is randomly shuffled and divided into meta-training (\cref{tab:train-table}), meta-validation (\cref{tab:val-table}), and meta-testing (\cref{tab:test-table}) sets, with a 75\%-10\%-15\% split, respectively. 
Dogs datasets for dog clade (group of breeds) prediction are used in meta-training, British humans datasets from the UK Biobank (UKB) for phenotype prediction are used in meta-validation, and HapMap3 datasets for subpopulation prediction are used in the meta-test. This strict separation ensures we meta-learn and evaluate on substantially different distributions and tasks.

\begin{table*}
  \centering
\begin{small}
  \begin{tabular}{lrrrrr}
\toprule
                          Dataset name &  Train size &  Test size &  Feature size &  Categorical &  Classes \\
\midrule
                   dogs$_{1..30}$ (30) &        1372 &        458 &           784 &          784 &       10 \\
                                  sick &        3017 &        755 &            29 &           22 &        2 \\
                           Bioresponse &        3000 &        751 &          1776 &            0 &        2 \\
                                splice &        2552 &        638 &            60 &           60 &        3 \\
                           qsar-biodeg &         844 &        211 &            41 &            0 &        2 \\
                           MiceProtein &         864 &        216 &            77 &            0 &        8 \\
                                isolet &        6237 &       1560 &           617 &            0 &       26 \\
                             connect-4 &       54045 &      13512 &            42 &           42 &        3 \\
                analcatdata\_authorship &         672 &        169 &            70 &            0 &        4 \\
                              kr-vs-kp &        2556 &        640 &            36 &           36 &        2 \\
                             optdigits &        4496 &       1124 &            64 &            0 &       10 \\
       analcatdata\_dmft &         637 &        160 &             4 &            4 &        6 \\
                                 churn &        4000 &       1000 &            20 &            4 &        2 \\
                        mfeat-karhunen &        1600 &        400 &            64 &            0 &       10 \\
                         mfeat-factors &        1600 &        400 &           216 &            0 &       10 \\
                                   kc1 &        1687 &        422 &            21 &            0 &        2 \\
                               texture &        4400 &       1100 &            40 &            0 &       11 \\
               Internet-Advertisements &        2623 &        656 &          1558 &         1555 &        2 \\
                                   har &        8239 &       2060 &           561 &            0 &        6 \\
jungle\_chess\_2pcs\_raw\_endgame\_complete &       35855 &       8964 &             6 &            0 &        3 \\
                                   car &        1382 &        346 &             6 &            6 &        4 \\
                              credit-g &         800 &        200 &            20 &           13 &        2 \\
                                 adult &       39073 &       9769 &            14 &            8 &        2 \\
                                 nomao &       27572 &       6893 &           118 &           29 &        2 \\
                                   jm1 &        8708 &       2177 &            21 &            0 &        2 \\
                           numerai28.6 &       77056 &      19264 &            21 &            0 &        2 \\
           first-order-theorem-proving &        4894 &       1224 &            51 &            0 &        6 \\
                                   dna &        2548 &        638 &           180 &          180 &        3 \\
                      Devnagari-Script &       73600 &      18400 &          1024 &            0 &       46 \\
                   mfeat-morphological &        1600 &        400 &             6 &            0 &       10 \\
                               madelon &        2080 &        520 &           500 &            0 &        2 \\
                                   pc3 &        1250 &        313 &            37 &            0 &        2 \\
      blood-transfusion-service-center &         598 &        150 &             4 &            0 &        2 \\
                               vehicle &         676 &        170 &            18 &            0 &        4 \\
                                 vowel &         792 &        198 &            12 &            2 &       11 \\
                         balance-scale &         500 &        125 &             4 &            0 &        3 \\
                segment &        1848 &        462 &            16 &            0 &        7 \\
                                   pc1 &         887 &        222 &            21 &            0 &        2 \\
                           tic-tac-toe &         766 &        192 &             9 &            9 &        2 \\
                               semeion &        1274 &        319 &           256 &            0 &       10 \\
                                letter &       16000 &       4000 &            16 &            0 &       26 \\
            electricity &       36249 &       9063 &             8 &            1 &        2 \\
     GesturePhaseSegmentationProcessed &        7898 &       1975 &            32 &            0 &        5 \\
                                cnae-9 &         864 &        216 &           856 &            0 &        9 \\
                       ozone-level-8hr &        2027 &        507 &            72 &            0 &        2 \\
                                  ilpd &         466 &        117 &            10 &            1 &        2 \\
  wall-robot-navigation &        4364 &       1092 &            24 &            0 &        4 \\
                         mfeat-fourier &        1600 &        400 &            76 &            0 &       10 \\
                              spambase &        3680 &        921 &            57 &            0 &        2 \\
                             mnist\_784 &       56000 &      14000 &           784 &            0 &       10 \\
                      PhishingWebsites &        8844 &       2211 &            30 &           30 &        2 \\
      climate-model-simulation-crashes &         432 &        108 &            18 &            0 &        2 \\
                    steel-plates-fault &        1552 &        389 &            27 &            0 &        7 \\
                           mfeat-pixel &        1600 &        400 &           240 &            0 &       10 \\
\bottomrule
\end{tabular}
\end{small}
  \caption{Meta-training datasets $\mathcal{D}_{\text{meta-train}}$. Train size is the number of training instances in $d_{\text{train}}$, and Test size is the number of test instances in $d_{\text{test}}$. Subscripts $i..j$ and $(\cdot)$ denote the interval of indices and the total number of datasets of the same group used, respectively.}
  \label{tab:train-table}
\end{table*}

\begin{table*}[!ht]
  \centering
  
\begin{small}
  \begin{tabular}{lrrrrr}
\toprule
  Dataset name &  Train size &  Test size &  Feature size &  Categorical &  Classes \\
\midrule
cylinder-bands &         432 &        108 &            37 &           19 &        2 \\
          wdbc &         455 &        114 &            30 &            0 &        2 \\
    eucalyptus &         588 &        148 &            19 &            5 &        5 \\
 mfeat-zernike &        1600 &        400 &            47 &            0 &       10 \\
           cmc &        1178 &        295 &             9 &            7 &        3 \\
 dresses-sales &         400 &        100 &            12 &           11 &        2 \\
      breast-w &         559 &        140 &             9 &            0 &        2 \\
      red hair &       24638 &       6160 &          1621 &         1621 &        2 \\
   blonde hair &       62297 &      15575 &          6968 &         6968 &        2 \\
dark brown hair &      202459 &      50615 &          5662 &         5662 &        2 \\
   black hair &       23001 &       5751 &          1649 &         1649 &        2 \\
     skin burn &       94972 &      23744 &          3158 &         3158 &        2 \\
      skin tan &      108592 &      27148 &          4130 &         4130 &        2 \\
  brown hair &      114502 &      28626 &          4024 &         4024 &        2 \\
malabsorption-coeliac &        3672 &        918 &           423 &          423 &        2 \\
\bottomrule
\end{tabular}
\end{small}
  \caption{Meta-validation datasets $\mathcal{D}_{\text{meta-val}}$. Train size is the number of training instances in $d_{\text{train}}$, and Test size is the number of test instances in $d_{\text{test}}$.}
  \label{tab:val-table}
\end{table*}

\begin{table*}[!ht]
  \centering
\begin{small}
\begin{tabular}{lrrrrr}
\toprule
           Dataset name &  Train size &  Test size &  Feature size &  Categorical &  Classes \\
\midrule
 hapmap$_{1..5}$ (5) &        1660 &        554 &           784 &          784 &       10 \\
                phoneme &        4323 &       1081 &             5 &            0 &        2 \\
                   wilt &        3871 &        968 &             5 &            0 &        2 \\
              pendigits &        8793 &       2199 &            16 &            0 &       10 \\
               satimage &        5144 &       1286 &            36 &            0 &        6 \\
        credit-approval &         552 &        138 &            15 &            9 &        2 \\
banknote-authentication &        1097 &        275 &             4 &            0 &        2 \\
         bank-marketing &       36168 &       9043 &            16 &            9 &        2 \\
                    pc4 &        1166 &        292 &            37 &            0 &        2 \\
                    kc2 &         417 &        105 &            21 &            0 &        2 \\
               diabetes &         614 &        154 &             8 &            0 &        2 \\
\bottomrule
\end{tabular}
\end{small}
  \caption{Meta-testing datasets $\mathcal{D}_{\text{meta-test}}$. Train size is the number of training instances in $d_{\text{train}}$, and Test size is the number of test instances in $d_{\text{test}}$. Subscripts $i..j$ and $(\cdot)$ denote the interval of indices and the total number of datasets of the same group used, respectively.}
  \label{tab:test-table}
\end{table*}

\subsection{HyperFast and Baselines Implementation}

\subsubsection{HyperFast Training Details}

In the meta-training stage, HyperFast weights are learnt by 
generating the weights of a smaller model that solves a different training task $t \in \mathcal{T}_{\text{meta-train}}$ at each training step. $t$ is derived from a randomly selected dataset $d$ from the collection of meta-training datasets $\mathcal{D}_{\text{meta-train}}$. However, the gradient signal is too noisy for weight updates at every training step. We fix this issue by accumulating gradients across different tasks before performing an optimization step. We experiment with gradient accumulation of 2, 3, 5, 10, 25, 50, and 100 steps. In our experiments we find that, in general, a larger number of accumulation steps always yields a more stable loss curve. That is, the meta-model learns better from observing the variations across different datasets, rather than solving one task at a time. We use a total of 25 gradient accumulation steps, which already allows a stable training, without excessively prolonging convergence. Despite this, during meta-validation, we observe a tendency to overfit to the meta-training datasets over very long training times. We select the HyperFast model that achieves the best average performance across the meta-validation datasets. We also experimented with solving multiple tasks in a single pass, but it was not possible in many cases due to memory constraints. 
Another key architectural design choice that significantly stabilizes the training process is sharing the core parameters between hypernetwork modules. As a shared module we use 2 feed-forward layers with a hidden dimensionality of 1024 and ReLU activations. We also experimented with deeper shared modules and different architectures based on attention mechanisms and convolutions, however, training stability and model generalization were inferior. 
The HyperFast used in this work has 1.27 B parameters (4.7 GB of memory), which generates the weights of smaller models of 52.65 M parameters (200.8 MB). The model is trained for 100,000 steps with a learning rate of 0.0003 with the AdamW optimizer \cite{loshchilov2018decoupled}, which required 20 hours  on a single NVIDIA Tesla V100 SXM2 GPU.

\subsubsection{HyperFast Inference Details}

Once HyperFast is trained, the hypernetwork weights are frozen and HyperFast can be used as an off-the-shelf model to generate target networks. Significant improvements in performance can be achieved when selecting the optimal target model configuration for the task at hand by ensembling and fine-tuning the generated networks. In other words, the meta-model is fixed and ready to generate weights for a support set, without needing any hyperparameter tuning. 
For the fastest inference, predictions can be obtained by directly using the target network generated by HyperFast in a single forward pass. For slower but most accurate predictions, one can optimize the inference model configuration for each dataset by ensembling generated networks and fine-tuning them, using the recommended search space from \cref{tab:hf_search_space}.

\begin{table}[!ht]
\centering
\resizebox{\columnwidth}{!}{%
  \begin{tabular}{ll}
\toprule
Parameter & Range \\
\midrule
n\_ensemble & [1, 4, 8, 16, 32] \\
batch\_size & [1024, 2048] \\
nn\_bias & [True, False] \\
optimization & [None, ``optimize'', ``ensemble\_optimize''] \\
optimize\_steps & [1, 4, 8, 16, 32, 64, 128] \\
seed & [0, 1, ..., 9] \\
\bottomrule
\end{tabular}
}\caption{Recommended search space for the inference framework of HyperFast.}
\label{tab:hf_search_space}
\end{table}

\subsubsection{Baselines Hyperparameter Selection}

\begin{table*}[!ht]
  \centering
  \begin{tabular}{llll}
\toprule
Model & Hyperparameter & Sampling & Range \\
\midrule
\multirow{1}{*}{KNN} & n\_neighbors & randint & [1, 16] \\
\midrule
\multirow{4}{*}{Log. Reg.} & penalty & choice & [l1, l2, none] \\
& max\_iter & randint & [50, 500] \\
& fit\_intercept & choice & [True, False] \\
& C & loguniform & [$e^{-5}$, 5] \\
\midrule
\multirow{10}{*}{XGBoost} & learning\_rate & loguniform & [$e^{-7}$, 1] \\
& max\_depth & randint & [1, 10] \\
& subsample & uniform & [0.2, 1] \\
& colsample\_bytree & uniform & [0.2, 1] \\
& colsample\_bylevel & uniform & [0.2, 1] \\
& min\_child\_weight & loguniform & [$e^{-16}$, $e^{5}$] \\
& alpha & loguniform & [$e^{-16}$, $e^{2}$] \\
& lambda & loguniform & [$e^{-16}$, $e^{2}$] \\
& gamma & loguniform & [$e^{-16}$, $e^{2}$] \\
& n\_estimators & randint & [100, 4000] \\
\midrule
\multirow{9}{*}{LightGBM} & num\_leaves & randint & [5, 50] \\
& max\_depth & randint & [3, 20] \\
& learning\_rate & loguniform & [$e^{-3}$, 1] \\
& n\_estimators & randint & [50, 2000] \\
& min\_child\_weight & loguniform & [$e^{-5}$, $e^{4}$] \\
& subsample & uniform & [0.2, 0.8] \\
& colsample\_bytree & uniform & [0.2, 0.8] \\
& reg\_alpha & choice & [0, 0.1, 1, 2, 5, 7, 10, 50, 100] \\
& reg\_lambda & choice & [0, 0.1, 1, 5, 10, 20, 50, 100] \\
\midrule
\multirow{6}{*}{CatBoost} & learning\_rate & loguniform & [$e^{-5}$, 1] \\
& random\_strength & randint & [1, 20] \\
& l2\_leaf\_reg & loguniform & [1, 10] \\
& bagging\_temperature & uniform & [0, 1] \\
& leaf\_estimation\_iterations & randint & [1, 20] \\
& iterations & randint & [100, 4000] \\
\midrule
\multirow{4}{*}{SAINT} & dim & choice & [32, 64, 128, 256] \\
& depth & choice & [1, 2, 3, 6, 12] \\
& heads & choice & [2, 4, 8] \\
& dropout & choice & [0, 0.1, 0.2, 0.3, 0.4, 0.5, 0.6, 0.7, 0.8] \\
\midrule
\multirow{5}{*}{MLP} & learning rate & loguniform & [$e^{-9}$, $e^{-3}$] \\
& batch size & uniform & [10, 2048] \\
& optimizer & choice & [Adam, AdamW, SGD, RMSprop] \\
& patience & uniform & [10, 50] \\
& validation split & uniform & [0.05, 0.5] \\
\midrule
\multirow{2}{*}{Net-DNF} & number of formulas & choice & [64, 128, 256, 512, 1024, 2048, 3072] \\
& feature selection beta & choice & [1.6, 1.3, 1., 0.7, 0.4, 0.1] \\
\bottomrule
\end{tabular}
  \caption{Hyperparameter search spaces for baseline methods. Hyperparameter configurations are drawn using the sampling technique specified in every range.}
  \label{tab:hyperparams}
\end{table*}

\begin{figure*}[!ht]
    \centering
    \includegraphics[width=\textwidth]{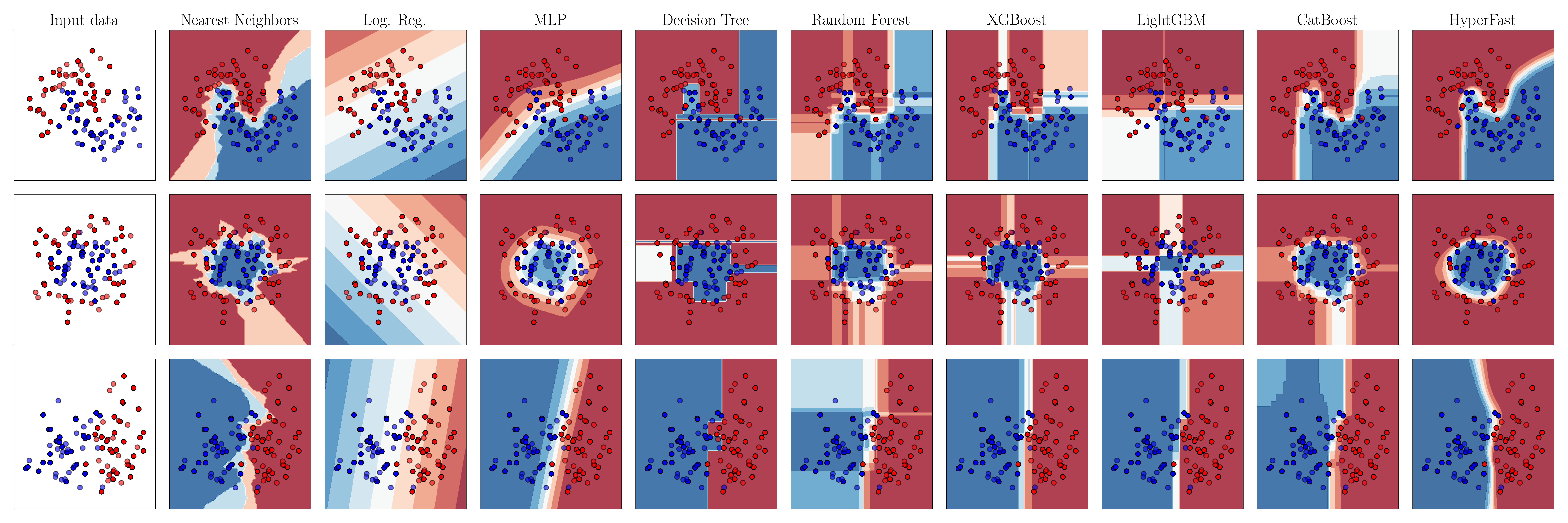}
    \caption{Classifiers comparison with the decision boundaries for toy binary classification datasets.}
    \label{fig:toy_datasets}
\end{figure*}

For hyperparameter tuning of the baselines, we use Hyperopt \cite{bergstra2015hyperopt}, a Python library for hyperparameter optimization through Bayesian optimization. For XGBoost and CatBoost we adapt the hyperparameter search spaces from \cite{shwartz2022tabular} and \cite{hollmann2023tabpfn}, which also tried other search spaces fixing the number of iterations and yielded suboptimal performance. For LightGBM we use the default hyperparameter search space defined in Hyperopt-sklearn \cite{komer2014hyperopt}. For KNN and Logistic Regression we use the ranges used in \cite{hollmann2023tabpfn}, while for SAINT, the search space implemented in \cite{borisov2021survey}. We benchmark DANet according to the configuration detailed in~\cite{chen2022danets}, and for Net-DNF \cite{katzir2020net}, we follow the search space suggested by the authors. For NODE \cite{popov2019neural}, FT-Transformer \cite{gorishniy2021revisiting}, and T2G-Former \cite{yan2023t2g}, we conducted experiments with the search spaces provided in the original papers. However, the computational and runtime costs associated with these methods is very high, making it impractical to thoroughly explore the search spaces within the 48-hour limit set for the evaluation corpus. In every instance, configurations explored below this limit yielded inferior results compared to the default configuration of each method, with the default implementations, surpassing the time limit on the big test. As an alternative to these challenges, we use the default configuration of the models on the mini test, and also perform a sweep of epochs until early stopping is performed (default case) to assess the improvement in performance within the runtime ranges of the models comparison. For the MLP, we replicate the exact same architecture as the main network produced by HyperFast, including the initial RF and PCA transformation layers. We perform hyperparameter tuning on the training hyperparameters, fixing the number of epochs to 1,000,000 and performing early stopping based on the validation loss. For TabPFN we consider up to 4096 data permutations for ensembling. \cref{tab:hyperparams} details the hyperparameter search spaces, as well as the sampling method used in every range for hyperparameter selection.

\begin{figure*}[!ht]
    \centering
    \includegraphics[width=\textwidth]{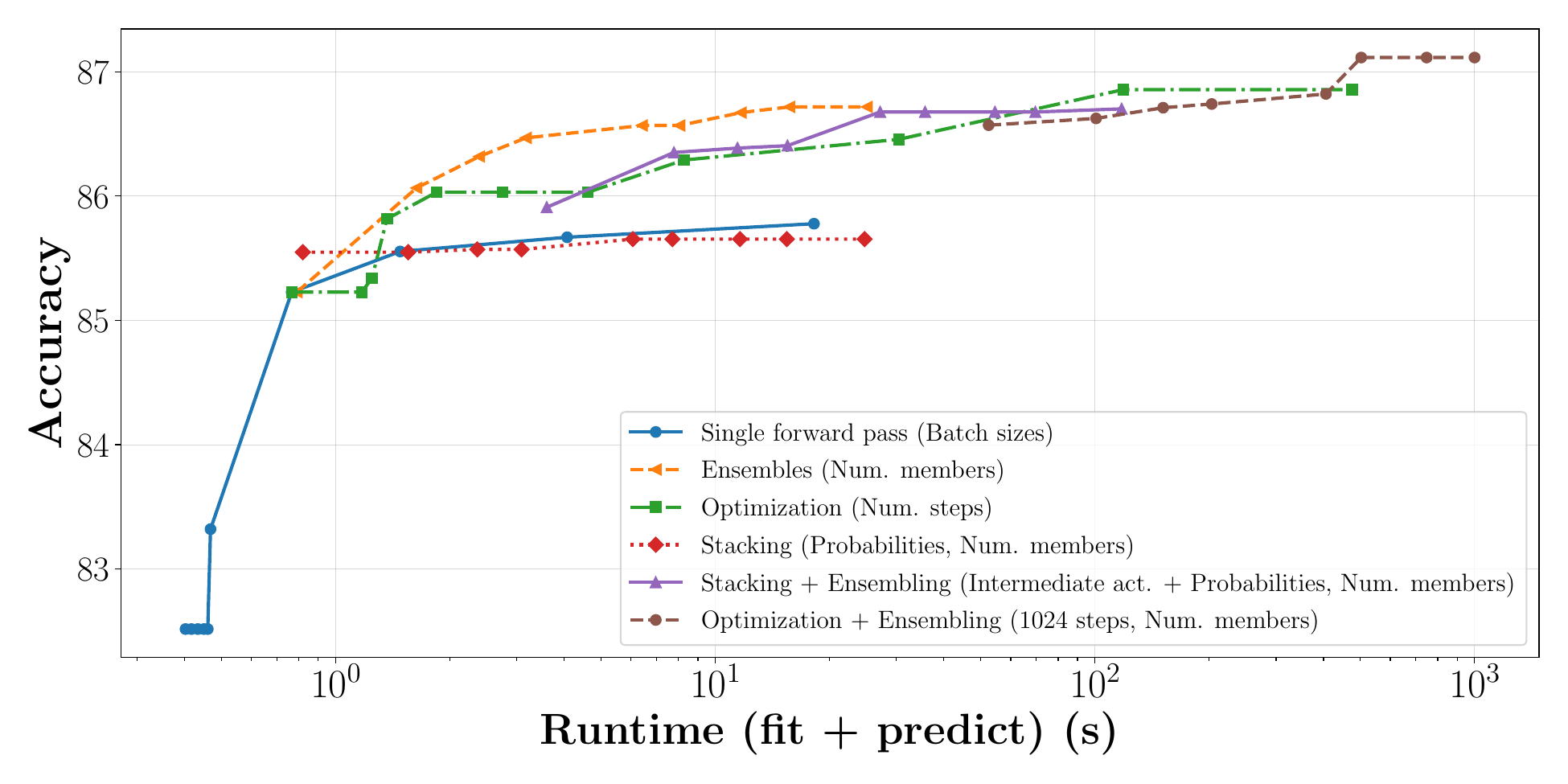}
    \caption{Performance as a function of runtime for different approaches to fully leverage all training data in the generation of the final inference model with HyperFast. Batch sizes considered in a single forward pass: [64, 128, 256, 512, 784, 1024, 2048, 4096, 8192, 16384]. Number of members considered in options involving ensembling or stacking: [1, 2, 3, 4, 8, 10, 15, 20, 32]. Optimization steps trials: [0, 2, 3, 4, 8, 16, 32, 64, 256, 1024, 4096].}
    \label{fig:leverage}
\end{figure*}

\begin{table*}[!ht]
  \centering
\begin{small}
\resizebox{\textwidth}{!}{%

 \begin{tabular}{lllllllllllll}
\toprule
 & Log. Reg. & XGBoost & LightGBM & CatBoost & MLP* & ASKL 2.0 & SAINT & DANet & Net-DNF & TabPFN & AutoGluon & HyperFast \\
\midrule
hapmap$_{1}$ & 47.899 $\pm$ 2.618 & 45.598 $\pm$ 2.412 & 46.037 $\pm$ 1.668 & 44.433 $\pm$ 1.555 & 46.870 $\pm$ 1.409 & 47.816 $\pm$ 1.928 & 39.750 $\pm$ 2.619 & 31.043 $\pm$ 8.923 & 39.288 $\pm$ 1.625 & 41.339 $\pm$ 1.942 & \bfseries 47.992 $\pm$ 2.664 & 47.758 $\pm$ 1.437 \\
hapmap$_{2}$ & 49.675 $\pm$ 2.149 & 45.754 $\pm$ 2.242 & 47.502 $\pm$ 2.175 & 46.654 $\pm$ 1.709 & 48.564 $\pm$ 1.324 & \bfseries 50.60 $\pm$ 1.924 & 40.534 $\pm$ 3.689 & 30.241 $\pm$ 5.934 & 39.776 $\pm$ 1.737 & 42.807 $\pm$ 1.650 & 49.598 $\pm$ 2.198 & 48.490 $\pm$ 1.736 \\
hapmap$_{3}$ & \bfseries 49.702 $\pm$ 1.422 & 45.862 $\pm$ 2.421 & 47.613 $\pm$ 2.777 & 45.028 $\pm$ 2.389 & 48.221 $\pm$ 1.713 & 49.142 $\pm$ 2.082 & 38.380 $\pm$ 2.648 & 29.952 $\pm$ 8.464 & 38.917 $\pm$ 1.602 & 41.109 $\pm$ 2.151 & 48.920 $\pm$ 2.044 & 48.261 $\pm$ 2.020 \\
hapmap$_{4}$ & \bfseries 50.331 $\pm$ 1.990 & 47.911 $\pm$ 2.854 & 46.988 $\pm$ 2.645 & 45.706 $\pm$ 3.018 & 48.473 $\pm$ 2.209 & 49.065 $\pm$ 2.190 & 40.792 $\pm$ 2.958 & 32.588 $\pm$ 6.857 & 39.580 $\pm$ 2.525 & 43.118 $\pm$ 1.945 & 50.077 $\pm$ 1.605 & 49.216 $\pm$ 2.082 \\
hapmap$_{5}$ & 50.221 $\pm$ 2.092 & 47.567 $\pm$ 3.638 & 48.385 $\pm$ 2.332 & 47.733 $\pm$ 2.917 & 49.761 $\pm$ 2.173 & \bfseries 50.884 $\pm$ 3.262 & 41.271 $\pm$ 3.099 & 36.132 $\pm$ 7.936 & 39.925 $\pm$ 2.461 & 42.011 $\pm$ 3.045 & 50.668 $\pm$ 2.50 & 50.173 $\pm$ 2.156 \\
phoneme & 65.172 $\pm$ 3.023 & 80.824 $\pm$ 1.631 & 80.968 $\pm$ 1.513 & \bfseries 81.926 $\pm$ 1.634 & 79.780 $\pm$ 1.331 & 81.831 $\pm$ 1.542 & 79.548 $\pm$ 1.154 & 75.092 $\pm$ 4.147 & 50.0 $\pm$ 0.0 & 81.080 $\pm$ 1.274 & 81.748 $\pm$ 1.128 & 80.794 $\pm$ 0.887 \\
wilt & 69.774 $\pm$ 7.629 & 85.155 $\pm$ 3.311 & 85.535 $\pm$ 2.320 & 85.957 $\pm$ 2.979 & 91.225 $\pm$ 2.727 & 88.956 $\pm$ 1.718 & 50.0 $\pm$ 0.0 & 50.0 $\pm$ 0.0 & 50.0 $\pm$ 0.0 & \bfseries 91.420 $\pm$ 3.576 & 87.351 $\pm$ 3.730 & 89.054 $\pm$ 4.225 \\
pendigits & 92.974 $\pm$ 0.594 & 96.348 $\pm$ 0.847 & 96.805 $\pm$ 0.484 & 97.714 $\pm$ 0.385 & 97.936 $\pm$ 0.245 & 97.433 $\pm$ 0.447 & 77.179 $\pm$ 2.048 & 96.152 $\pm$ 2.768 & 80.425 $\pm$ 3.754 & \bfseries 98.657 $\pm$ 0.241 & 97.783 $\pm$ 0.276 & 98.498 $\pm$ 0.378 \\
satimage & 79.126 $\pm$ 0.955 & 85.451 $\pm$ 1.025 & 85.654 $\pm$ 0.705 & 85.481 $\pm$ 0.748 & 85.372 $\pm$ 1.647 & 85.736 $\pm$ 0.949 & 84.739 $\pm$ 1.713 & 79.597 $\pm$ 8.511 & 79.048 $\pm$ 1.482 & 85.361 $\pm$ 1.066 & 85.984 $\pm$ 0.853 & \bfseries 86.437 $\pm$ 0.542 \\
credit-appr. & 84.352 $\pm$ 0.0 & \bfseries 87.290 $\pm$ 0.0 & 87.0 $\pm$ 0.082 & 84.352 $\pm$ 0.0 & 84.059 $\pm$ 0.850 & 84.257 $\pm$ 0.708 & 83.334 $\pm$ 0.891 & 80.892 $\pm$ 1.836 & 80.777 $\pm$ 1.264 & 80.594 $\pm$ 0.0 & 84.694 $\pm$ 0.987 & 80.594 $\pm$ 0.0 \\
bank.-auth. & 98.693 $\pm$ 0.0 & 99.739 $\pm$ 0.138 & 99.837 $\pm$ 0.172 & 99.673 $\pm$ 0.0 & \bfseries 100.0 $\pm$ 0.0 & \bfseries 100.0 $\pm$ 0.0 & \bfseries 100.0 $\pm$ 0.0 & 98.733 $\pm$ 1.519 & 93.727 $\pm$ 2.938 & \bfseries 100.0 $\pm$ 0.0 & \bfseries 100.0 $\pm$ 0.0 & \bfseries 100.0 $\pm$ 0.0 \\
bank-mkt. & 65.952 $\pm$ 2.422 & 64.973 $\pm$ 2.773 & 64.507 $\pm$ 2.980 & 65.928 $\pm$ 2.215 & 59.738 $\pm$ 3.748 & 62.806 $\pm$ 5.105 & 57.260 $\pm$ 4.854 & 50.376 $\pm$ 1.180 & 51.802 $\pm$ 2.549 & 61.591 $\pm$ 2.699 & 61.054 $\pm$ 3.546 & \bfseries 76.669 $\pm$ 1.330 \\
pc4 & 68.351 $\pm$ 0.210 & 75.161 $\pm$ 1.205 & 74.963 $\pm$ 1.554 & 73.099 $\pm$ 1.918 & 71.497 $\pm$ 1.979 & 74.737 $\pm$ 3.870 & 66.743 $\pm$ 4.820 & 50.0 $\pm$ 0.0 & 59.314 $\pm$ 6.702 & 72.170 $\pm$ 1.249 & 71.359 $\pm$ 2.406 & \bfseries 75.662 $\pm$ 2.736 \\
kc2 & 65.170 $\pm$ 0.0 & 67.908 $\pm$ 0.0 & 67.442 $\pm$ 0.0 & 63.946 $\pm$ 1.761 & 66.265 $\pm$ 2.503 & 64.540 $\pm$ 0.758 & 65.115 $\pm$ 2.194 & 63.245 $\pm$ 5.950 & 50.0 $\pm$ 0.0 & \bfseries 69.113 $\pm$ 0.0 & 62.878 $\pm$ 1.905 & 67.442 $\pm$ 0.0 \\
diabetes & 66.926 $\pm$ 0.0 & \bfseries 72.204 $\pm$ 0.0 & 71.280 $\pm$ 0.327 & 69.630 $\pm$ 1.263 & 70.280 $\pm$ 1.161 & 67.593 $\pm$ 2.703 & 58.713 $\pm$ 3.241 & 63.261 $\pm$ 5.322 & 58.565 $\pm$ 6.963 & 68.778 $\pm$ 0.0 & 68.935 $\pm$ 1.431 & 70.907 $\pm$ 0.0 \\
\midrule Mean rank & 5.953 $\pm$ 0.191 & 4.747 $\pm$ 0.514 & 4.660 $\pm$ 0.488 & 5.347 $\pm$ 0.514 & 4.720 $\pm$ 0.478 & 4.30 $\pm$ 0.542 & 8.533 $\pm$ 0.439 & 9.373 $\pm$ 0.524 & 10.113 $\pm$ 0.252 & 6.093 $\pm$ 0.308 & 4.273 $\pm$ 0.341 & \bfseries 3.547 $\pm$ 0.408 \\
\midrule Mean bal. acc. & 66.955 $\pm$ 0.525 & 69.850 $\pm$ 0.316 & 70.034 $\pm$ 0.461 & 69.151 $\pm$ 0.328 & 69.869 $\pm$ 0.372 & 70.360 $\pm$ 0.492 & 61.557 $\pm$ 0.647 & 57.820 $\pm$ 1.470 & 56.743 $\pm$ 0.827 & 67.943 $\pm$ 0.435 & 69.936 $\pm$ 0.455 & \bfseries 71.330 $\pm$ 0.445 \\
\bottomrule
\end{tabular}
}\end{small}
  \caption{Balanced accuracy results per dataset on the mini test for a runtime budget of 5 minutes. The mean rank of each method is also shown, for 10 repetitions with different selection of samples and features to subset and create the mini test. MLP*: MLP with the exact same architecture as the main network produced by HyperFast, including the initial RF and PCA transformation layers.}
  \label{tab:table_mini_test_5min}
\end{table*}

\begin{table*}[!ht]
  \centering
\begin{small}
\resizebox{\textwidth}{!}{%
 \begin{tabular}{llllllllllllllll}
\toprule
 & LR & XGB & LGBM & CatB & MLP* & ASKL2 & SAINT & DANet & Net-DNF & NODE & TabPFN & FT-T & T2G & AG & HF \\
\midrule
hapmap$_{1}$ & 48.195 & 45.353 & 46.649 & 47.209 & 47.490 & 45.278 & 47.049 & 19.868 & 43.832 & 43.303 & 41.262 & 44.304 & 43.787 & 47.769 & \bfseries 49.640 \\
hapmap$_{2}$ & 51.434 & 51.372 & 47.968 & 48.139 & 51.529 & 49.552 & 49.332 & 27.122 & 48.572 & 49.535 & 42.778 & 50.242 & 48.085 & \bfseries 51.741 & 51.197 \\
hapmap$_{3}$ & 50.294 & 47.162 & 48.821 & 44.927 & 48.348 & 50.087 & \bfseries 50.384 & 28.614 & 44.549 & 47.806 & 40.464 & 49.230 & 47.070 & 49.511 & 49.701 \\
hapmap$_{4}$ & 51.347 & 49.222 & 50.676 & 47.784 & 49.808 & 51.299 & 46.988 & 31.779 & 44.880 & 45.332 & 40.572 & 45.256 & 47.604 & \bfseries 53.314 & 49.640 \\
hapmap$_{5}$ & \bfseries 52.170 & 52.122 & 50.934 & 51.282 & 52.133 & 49.512 & 50.058 & 28.311 & 47.017 & 46.791 & 41.189 & 48.647 & 47.195 & 51.131 & 51.190 \\
phoneme & 66.526 & 82.420 & 81.685 & 83.601 & 82.035 & \bfseries 83.769 & 80.716 & 70.754 & 50.0 & 80.403 & 80.703 & 82.953 & 81.691 & 82.970 & 81.146 \\
wilt & 77.284 & 89.259 & 90.003 & 90.166 & 91.019 & 91.073 & 50.0 & 50.0 & 50.0 & 50.0 & 88.352 & 92.833 & \bfseries 95.587 & 90.964 & 91.073 \\
pendigits & 93.944 & 97.038 & 97.155 & 98.085 & 97.978 & 98.062 & 76.753 & 97.897 & 90.571 & 96.250 & \bfseries 98.823 & 97.486 & 97.541 & 98.291 & 98.652 \\
satimage & 80.791 & 86.493 & 86.721 & 86.857 & 86.317 & 87.155 & 86.420 & 84.769 & 83.480 & 87.016 & 87.358 & 86.073 & 86.274 & 87.074 & \bfseries 87.416 \\
credit-appr. & 84.352 & 86.30 & \bfseries 87.119 & 86.30 & 85.650 & 85.480 & 84.831 & 78.720 & 83.532 & 84.288 & 81.893 & 86.438 & 86.268 & 85.650 & 82.063 \\
bank.-auth. & 98.693 & \bfseries 100.0 & \bfseries 100.0 & 99.673 & \bfseries 100.0 & \bfseries 100.0 & \bfseries 100.0 & 98.854 & 99.673 & \bfseries 100.0 & \bfseries 100.0 & \bfseries 100.0 & \bfseries 100.0 & \bfseries 100.0 & \bfseries 100.0 \\
bank-mkt. & 70.475 & 69.063 & 67.949 & 70.859 & 70.508 & 62.490 & 61.104 & 56.968 & 54.101 & 64.596 & 65.0 & 70.648 & 64.214 & 65.529 & \bfseries 75.656 \\
pc4 & 68.663 & 78.798 & 77.604 & 76.411 & 76.432 & \bfseries 83.181 & 72.873 & 49.609 & 61.914 & 66.102 & 74.631 & 70.877 & 66.710 & 73.438 & 78.212 \\
kc2 & 65.170 & 64.567 & 67.442 & 64.567 & 69.113 & 67.908 & 66.840 & 61.829 & 65.772 & 65.170 & 69.715 & 69.113 & 66.375 & 65.635 & \bfseries 72.453 \\
diabetes & 67.0 & 75.759 & 71.130 & 71.981 & 71.981 & \bfseries 76.889 & 63.444 & 67.074 & 71.278 & 68.352 & 70.204 & 68.074 & 64.944 & 70.556 & 70.907 \\
\midrule Mean rank & 7.467 & 5.933 & 6.20 & 5.667 & 4.533 & 4.533 & 8.60 & 12.467 & 11.067 & 9.733 & 8.80 & 6.667 & 8.667 & 4.60 & \bfseries 3.933 \\
\midrule Mean bal. acc. & 68.423 & 71.662 & 71.457 & 71.189 & 72.023 & 72.116 & 65.786 & 56.811 & 62.611 & 66.330 & 68.196 & 70.812 & 69.556 & 71.572 & \bfseries 72.596 \\
\bottomrule
\end{tabular}
}
\end{small}
  \caption{Balanced accuracy results per dataset on the mini test with extended runtime. The mean rank of each method is also shown. LR: Logistic Regression; XGB: XGBoost; LGBM: LightGBM; CatB: CatBoost; MLP*: MLP with the exact same architecture as the main network produced by HyperFast, including the initial RF and PCA transformation layers; ASKL2: ASKL 2.0; FT-T: FT-Transformer; T2G: T2G-Former; AG: AutoGluon; HF: HyperFast. }
  \label{tab:table_mini_test}
\end{table*}

\section{Additional Results}

\subsection{Toy datasets}

In \cref{fig:toy_datasets} we compare HyperFast to traditional ML methods on toy datasets from scikit-learn \cite{scikit-learn}: \emph{make\_moons} in the top row, \emph{make\_circles} in the middle row, and a linearly separable dataset in the bottom row, all with Gaussian noise added. We can see how HyperFast models correctly the \emph{moons} and \emph{circles} without overfitting to the outliers, and creates a reasonably linear decision boundary for the bottom case. In contrast, tree-based methods overfit to the training data and fail to model accurately the distributions, creating abrupt and inaccurate decision boundaries in most cases.

\subsection{How Can We Leverage All Labeled Data of a Large Dataset?}
\label{sec:leverage_data}

In a single forward pass, HyperFast can generate a set of weights for a smaller model ready for inference using a set of labeled samples.
However, for datasets with large training sets it is not possible to use all available labeled data in a single forward pass due to memory and efficiency constraints, thus possibly losing relevant information from the dataset that could be valuable for the generation of weights to solve the task. We compare different options to leverage all labeled data in the generation of the final inference model in \cref{fig:leverage}.

We first experiment with increasing the batch size in a single forward pass. As we can expect, larger batch sizes yield significantly better performance, but at the cost of a much slower runtime. This is mainly due to the singular value decomposition (SVD) performed in the PCA module, although implemented and optimized for GPU, the computation time scales rapidly with the number of input samples when an excessively large batch size is used. 
Thus, for the trained HyperFast and for the rest of experiments, we use a fixed maximum batch size of 2048 samples, which yields very good results in less than a second.

Multiple models can be generated from different subsets of datapoints, each capturing different variations between samples. Additionally, the random features projection matrix is reinitialized in every forward pass of HyperFast, injecting more variability in all the following generated layers across models, even if the same subset of samples is used in different forward passes. We combine the predictions of multiple generated models with soft-voting ensembles, and we observe that bigger ensembles make more accurate predictions. 
Another alternative we experiment with is stacking the predictions of multiple main models using a Logistic Regression as the meta-learner. However, performance stagnates and does not improve with more stacking members. 
We also try a variant of stacking, where instead of stacking predictions from multiple models and fitting a single meta-learner, we stack the predictions and all intermediate activations from a single model and fit a meta-learner. We repeat the process for several main models and meta-learners, creating an ensemble of meta-learners. Although it is a more expensive process, we find that it yields better results than traditional stacking, performing on par with ensembling but with higher runtimes. 
Furthermore, we consider the weights generated by HyperFast as an starting point for fine-tuning the model on all training data. Note that in this case, all model weights are optimized: random features, PCA parameters, and linear layer weights. In \cref{fig:leverage} we see that optimizing the generated model in a single forward pass with all the training data, results are worse than ensembling for a small runtime budget. But for larger runtimes, optimization outperforms ensembling and stacking on their own. 
Finally, we combine the two fastest and best performing options, i.e., \emph{Optimization + Ensembling}, where we generate models in different forward passes, optimize them, and combine the fine-tuned models by ensembling. We perform 1024 fine-tuning steps in each generated network with a batch size of 2048, using the AdamW optimizer with a learning rate of 1e-4, and a scheduler that reduces the learning rate by a factor of 0.1 when the loss stagnates for 10 steps. We observe that although this combination requires more runtime, a single fine-tuned model matches the performance of large ensembles of non-optimized models, and a large ensemble of fine-tuned models yields the best results. 
We show results by starting with a single forward pass, then increasing the ensemble size by performing multiple forward passes until GPU memory is overloaded. Then, we restart the sweep by optimizing each generated model and ensembling the fine-tuned networks.

\subsection{Extended Results of Experiments}
\label{sec:extended_results}

\begin{figure*}[!t]
    \centering
    \includegraphics[width=\textwidth]{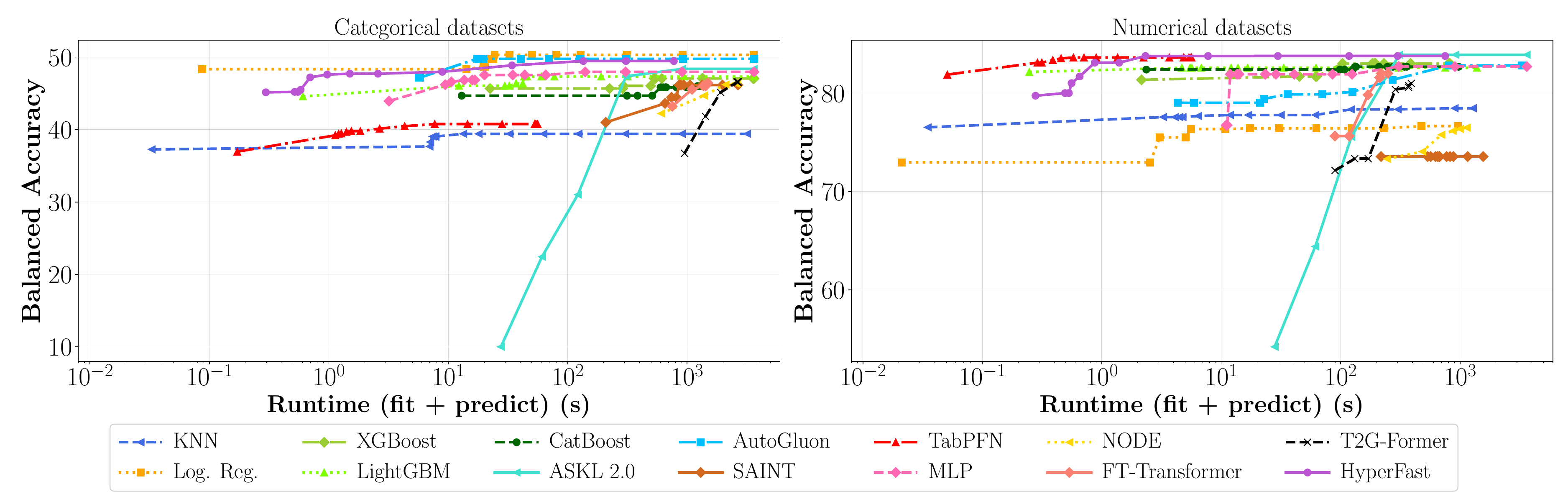} 
  \caption{(left) Categorical datasets of the mini test. (right) Numerical datasets of the mini test.}
    \label{fig:categorical_numerical_mini}
\end{figure*}

  \begin{figure*}[!ht]
  \centering
  \includegraphics[width=\textwidth]{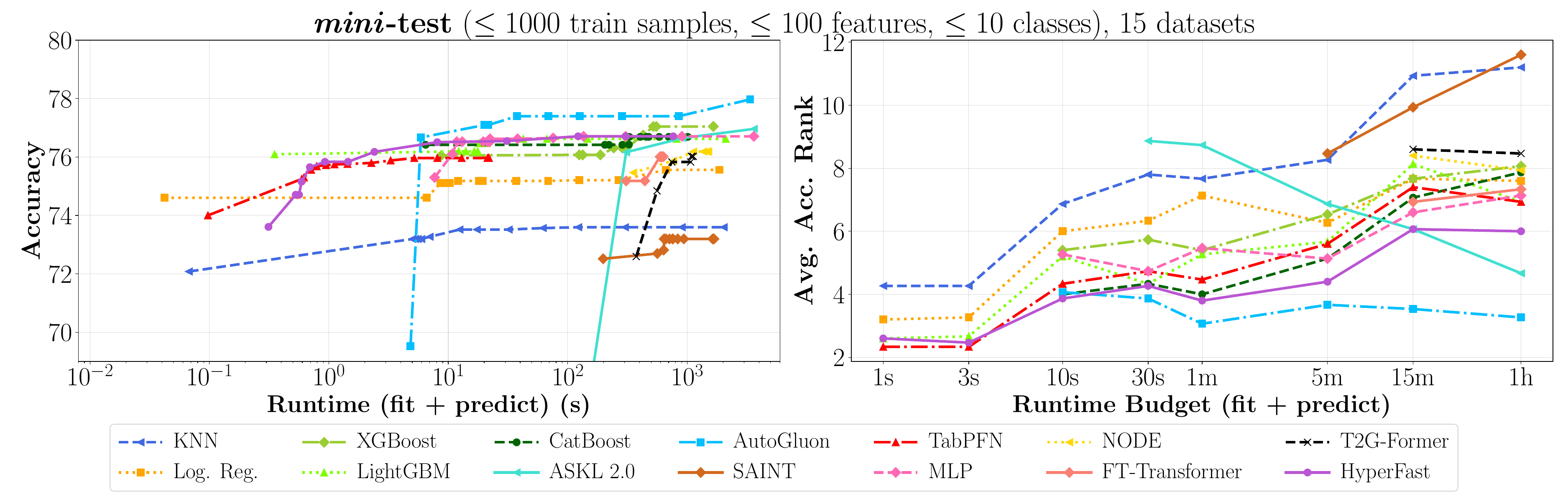} 
  \caption{Runtime (fit + predict) vs. regular accuracy and average rank for given runtime budgets on the mini test: 15 small-sized meta-datasets.}
  \label{fig:mini_test_plots_acc}
\end{figure*}

\begin{table*}[!ht]
  \centering
\begin{small}
\resizebox{\textwidth}{!}{
 \begin{tabular}{llllllllllll}
\toprule
 & LR & XGB & LGBM & CatB & MLP* & ASKL2 & SAINT & DANet & Net-DNF & AG & HF \\
\midrule
hapmap$_{1}$ & 74.299 & 71.396 & 70.750 & 68.608 & 74.169 & 76.397 & 61.874 & 16.029 & 58.501 & 76.712 & \bfseries 80.497 \\
hapmap$_{2}$ & 73.507 & 63.447 & 70.622 & 70.695 & 73.509 & 75.869 & 62.326 & 29.873 & 63.359 & 77.210 & \bfseries 78.693 \\
hapmap$_{3}$ & 76.007 & 66.742 & 71.436 & 72.079 & 75.949 & 78.481 & 60.510 & 65.724 & 62.977 & \bfseries 79.548 & 78.828 \\
hapmap$_{4}$ & 75.60 & 74.043 & 71.346 & 68.860 & 75.275 & 77.388 & 52.228 & 66.506 & 62.718 & 79.894 & \bfseries 81.782 \\
hapmap$_{5}$ & 79.435 & 67.059 & 72.009 & 72.045 & 80.106 & 80.797 & 61.833 & 32.847 & 64.505 & 82.374 & \bfseries 83.364 \\
phoneme & 64.553 & 84.717 & 86.633 & 85.467 & 83.482 & \bfseries 87.975 & 81.384 & 70.323 & 81.116 & 86.422 & 83.863 \\
wilt & 69.974 & 89.150 & 88.189 & 91.019 & 91.980 & \bfseries 93.958 & 50.0 & 50.0 & 69.231 & 93.051 & 93.903 \\
pendigits & 94.737 & 98.998 & 99.184 & 99.275 & \bfseries 99.596 & 99.232 & 94.622 & 98.721 & 91.945 & 99.505 & 99.501 \\
satimage & 81.057 & 89.390 & 89.824 & 89.708 & 89.157 & 89.655 & 86.325 & 89.319 & 82.235 & \bfseries 90.942 & 90.813 \\
credit-appr. & 84.352 & 86.949 & \bfseries 87.119 & 86.30 & 84.490 & 84.831 & 81.105 & 80.424 & 81.754 & 85.480 & 82.063 \\
bank.-auth. & 98.693 & 99.673 & \bfseries 100.0 & 99.673 & \bfseries 100.0 & \bfseries 100.0 & 99.673 & 99.673 & 93.453 & \bfseries 100.0 & \bfseries 100.0 \\
bank-mkt. & 66.174 & 71.593 & 73.814 & 73.771 & 72.952 & 75.426 & 71.687 & 65.436 & 66.523 & 71.491 & \bfseries 77.019 \\
pc4 & 68.273 & 75.022 & 76.606 & 76.606 & 74.240 & \bfseries 82.986 & 55.360 & 50.0 & 66.688 & 72.049 & 80.599 \\
kc2 & 65.170 & 61.090 & 68.045 & 64.567 & 71.249 & 67.908 & 67.908 & 67.442 & 50.0 & 63.499 & \bfseries 72.453 \\
diabetes & 67.0 & 72.056 & 71.130 & 71.981 & 72.407 & \bfseries 74.185 & 50.0 & 65.426 & 67.352 & 69.852 & 73.185 \\
\midrule Mean rank & 6.867 & 5.867 & 4.667 & 5.0 & 4.333 & 2.733 & 8.667 & 9.0 & 9.0 & 3.467 & \bfseries 2.267 \\
\midrule Mean bal. acc. & 75.922 & 78.088 & 79.780 & 79.377 & 81.237 & 83.006 & 69.122 & 63.183 & 70.824 & 81.868 & \bfseries 83.771 \\
\bottomrule
\end{tabular}
}
\end{small}
  \caption{Balanced accuracy results per dataset on the big test with extended runtime. The mean rank of each method is also shown. LR: Logistic Regression; XGB: XGBoost; LGBM: LightGBM; CatB: CatBoost; MLP*: MLP with the exact same architecture as the main network produced by HyperFast, including the initial RF and PCA transformation layers; ASKL2: ASKL 2.0; AG: AutoGluon; HF: HyperFast.
  }
  \label{tab:table_big_test}
\end{table*}

\begin{figure*}[!t]
    \centering
    \includegraphics[width=\textwidth]{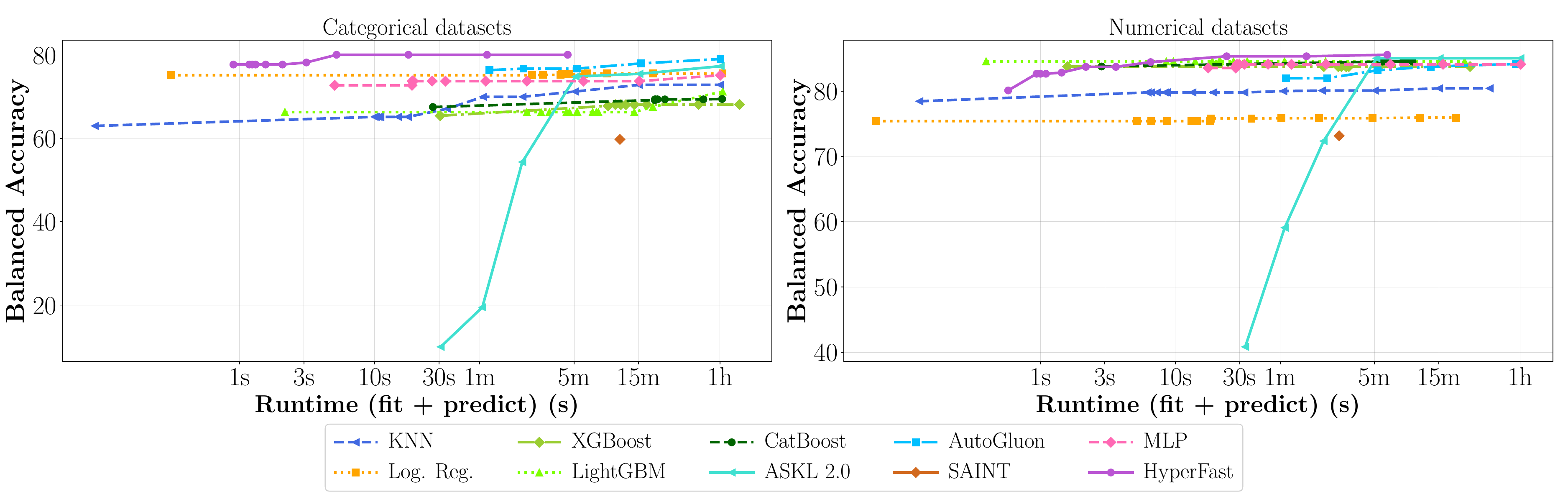} 
  \caption{(left) Categorical datasets of the big test. (right) Numerical datasets of the big test.}
    \label{fig:categorical_numerical_big}
\end{figure*}

 \begin{figure*}[!ht]
  \centering
  \includegraphics[width=\textwidth]{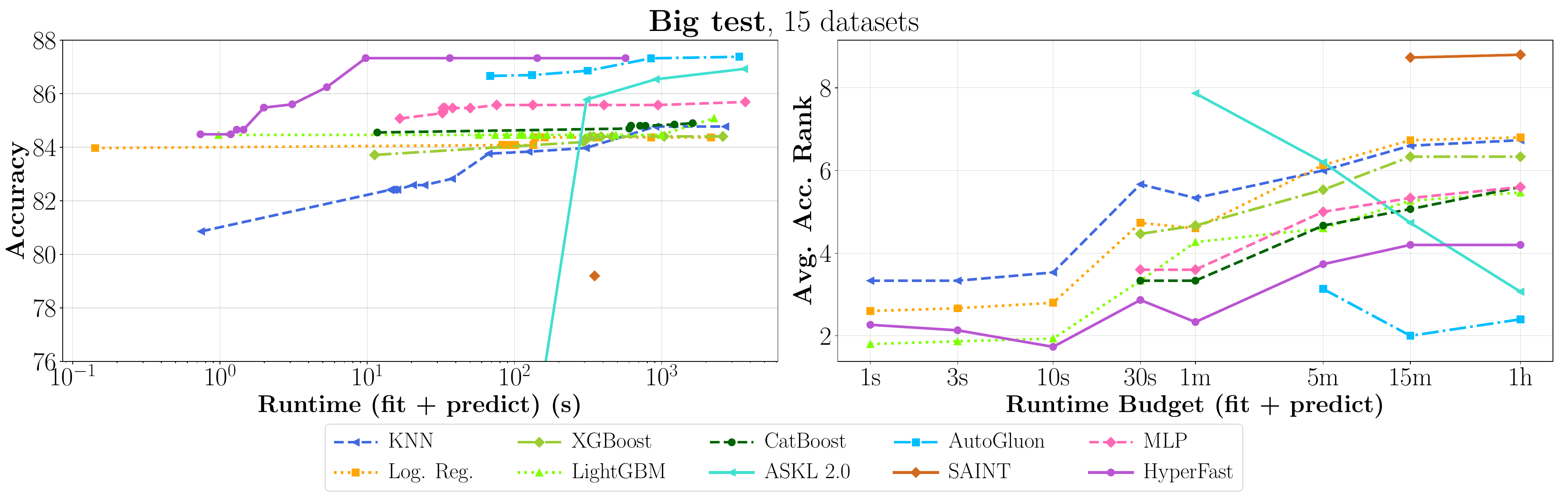} 
  \caption{Runtime (fit + predict) vs. regular accuracy and average rank for given runtime budgets on the big test: 15 large/medium-sized meta-datasets.}
  \label{fig:big_test_plots_acc}
\end{figure*}

Extending the results of the main paper, \cref{tab:table_mini_test_5min} shows per dataset results on the mini test, for a total runtime budget of 5 minutes, and 10 repetitions for different sample and feature subsetting to create the small-sized mini test. These results show that HyperFast is the best option for a rapid deployment setting, outperforming TabPFN, AutoML systems and other methods.
Additionally, \cref{tab:table_mini_test} shows the results on the mini test for a 1h budget, but we allow an extended total runtime of 48h on the 15 datasets. With a sufficient amount of time for hyperparameter tuning on such small datasets, HyperFast is the best overall performing method, followed by AutoML systems and the MLP matching the architecture of HyperFast's generated main network, including the RF+PCA initial layers. In \cref{fig:categorical_numerical_mini} we can see that TabPFN underperforms for categorical datasets but obtains competitive performance for numerical datasets in a very low runtime. However, it is outperformed by HyperFast with a runtime of 2 seconds. \cref{fig:mini_test_plots_acc} shows the mini-test results in terms of regular accuracy, where TabPFN surpasses HyperFast for low runtimes, and AutoGluon also obtains better accuracy for high runtimes. This is contrary to the balanced accuracy results, which indicates that these models may underperform in accurately predicting minority classes on imbalanced datasets.

For the large datasets setting, \cref{fig:categorical_numerical_big} shows the results of the different classifiers separated for fully categorical and numerical datasets. HyperFast obtains the best balanced accuracy results across all runtime regimes for categorical datasets. In contrast, gradient-boosting machines obtain better results for small time budgets on numerical datasets, but AutoML systems and HyperFast rapidly match and surpass their performance when more time is given to create larger ensembles and fine-tune each member. 
We show detailed results per dataset in \cref{tab:table_big_test} for the big test on a 1h budget per dataset, with a total extended runtime limit of 48h.
In a large-scale setting, tree-based gradient-boosting machines and the MLP are outperformed by AutoML systems which, in fact, train multiple instances of these gradient boosting algorithms and neural networks (among other models) to build a stronger predictor. The resulting ensemble is increasingly powerful when long fitting time budgets are allowed. 
\cref{fig:big_test_plots_acc} shows that HyperFast is outperformed by AutoML systems for long runtimes in terms of regular accuracy. However, HyperFast obtains the best balanced accuracy results, which suggests that HyperFast is the model that performs more consistently across different classes, particularly in datasets where some classes are underrepresented.

When it comes to individual datasets, HyperFast outperforms other methods, especially in fully categorical datasets. One reason is the use of PCA projections before the neural layers, and the concatenation of the global average and per class average of the PCA projections to each hypernetwork module. Previous work on genetic datasets~\cite{novembre2008genes} demonstrated the capability of PCA-based methods to capture the variation of samples and structure of the data. In the case of more diverse tabular datasets that have no underlying structure, the use of PCA does not have a negative impact. As a result, we have observed HyperFast also outperforming other baselines in diverse OpenML tabular datasets. When using a large number of principal components (PCs) (784) there is no information loss for datasets with $d$ features if $d\leq784$, which is the case for most datasets considered. Information loss in datasets with $d>784$ is minimal, since we keep the first 784 PCs associated with the largest eigenvalues, while the remaining components explain the least amount of variance in the data. The ablation studies show that even decreasing the number of PCs to 512, performance is not very affected, while removing the PCA transformation results in the largest drop in performance. 
The volume of support samples also has an effect on HyperFast's performance, as larger datasets provide more robust statistical basis for the hypernetwork to accurately predict weights. In contrast, a limited number of support samples may restrict the hypernetwork's ability to capture the statistical properties of the dataset, consequently affecting the accuracy of the generated main network.

\subsection{High-Dimensional Biomedical Datasets}
\label{ssec:high_dim_biomed}

In real-world biomedical applications, many tabular datasets exhibit very high dimensionality, making gradient-boosted trees computationally expensive and prone to overfitting, while traditional linear methods fail to capture non-linear interactions, leading to subpoptimal modeling performance. Additionally, current deep learning methods can struggle with scalability and present unfeasible training challenges when applied to datasets of such scale. 

A meta-trained and scalable model, such as HyperFast, offers a new approach to address these issues and provides an alternative classification approach for real-world applications. In this work, we conduct additional experiments on two high-dimensional biomedical datasets. 
First, we use \emph{hapmap-100k}, a HapMap3 \cite{international2010integrating} dataset following the steps described in the Experimental Setup, but randomly selecting a total of 100,000 SNPs from all the available SNPs without missing data. Next, we utilize \emph{diabetes-31k}, a UK Biobank diabetes prediction dataset including underrepresented populations~\cite{bonet2024machine}. While such biobanks include more diverse genetic backgrounds, the majority group in the UK Biobank includes individuals with European (British) ancestry, and other groups are still highly underrepresented. This dataset includes 31,153 SNPs for 66,302 individuals with European, South Asian, African, and East Asian ancestry. 

\begin{table}[!ht]
\centering
\begin{tabular}{lccc}
\toprule
\multirow{2}{*}{Model} & \multirow{2}{*}{hapmap-100k} & \multicolumn{2}{c}{diabetes-31k} \\
\cmidrule(lr){3-4}
 &  & All & Underrep. only \\
\midrule
Lasso & 93.564 & 54.365 & 53.376\\
Elastic Net & 94.208 & 53.454 & 52.708\\
LightGBM & 83.301 & 50.412 & 50.419\\
XGBoost & 82.475 & 50.561 & 50.351\\
HyperFast & \bfseries 95.889 & \bfseries 64.327 & \bfseries 54.023\\
\bottomrule
\end{tabular}
\caption{Balanced accuracy on high-dimensional biomedical datasets.}
\label{table:high_dim}
\end{table}

In the case of high-dimensional datasets where the support set including all features does not fit in GPU memory, we adapt HyperFast to perform feature bagging. For these experiments, we create ensembles of 32 networks and perform fine-tuning. For each ensemble member, HyperFast samples a subset of 3,000 SNPs from a multinomial distribution weighted by their standard deviation.

The results in \cref{table:high_dim} highlight HyperFast's robustness in high dimensional settings, achieving the best performance in both biomedical datasets, followed by the linear models. Remarkably, HyperFast obtains a 10\% increase in balanced accuracy in \emph{diabetes-31k} for test samples of all populations, and also obtains stronger predictions when only analyzing the performance for test individuals that are underrepresented in the training data.

\section{Limitations}

In terms of number of samples, HyperFast takes a fixed number of training samples (support set) to predict a single set of weights. For very large datasets, the generated main network in a single forward pass will not deliver optimal results, as the sample of data points used for the generation might not fully represent the entire dataset distribution. However, rapid improvements can be obtained with the optimization and ensembling techniques detailed previously, enabling the use of any dataset size. Regarding the number of features, the input size is not fixed, as HyperFast projects the original data of any given feature size with the random features and PCA module to a fixed size. In a single forward pass, the number of input features is only restricted by the amount of GPU memory available. To address this issue, feature bagging can be used together with ensembling for dealing with very high-dimensional datasets.
Note that if the number of selected features for an ensemble member is much larger than the number of PCs used, some information might be lost. To address this, one can train larger versions of HyperFast by increasing the number of retained PCs.

Our work prioritizes a simple yet effective method suitable for most tabular datasets within a constrained computational environment. Future work could explore expanding HyperFast to regression tasks and transitioning to a large-scale setup utilizing multiple GPUs for the meta-training of the model, where most information could be retained for very large numbers of features or different modalities (e.g., high resolution images). We also leave as future work the study of dataset distribution differences from meta-training and how it affects generalization performance. 
Lastly, in terms of number of categories, the HyperFast version discussed in this paper supports up to 100 classes. Nonetheless, training a HyperFast to accommodate more categories would increase linearly the complexity of the initial layers of the hypernetwork modules, which accounts for very small memory and computational requirements.  

\end{document}